\documentclass[10pt,journal,compsoc]{IEEEtran}
\ifCLASSOPTIONcompsoc
  \usepackage[nocompress]{cite}
\else
  \usepackage{cite}
\fi
\ifCLASSINFOpdf
\else
\fi

\usepackage{graphicx}
\usepackage{amsmath}
\usepackage{amssymb}
\usepackage{booktabs}
\usepackage{color}

\newcommand{\HF}[1]{{\color{black}#1}}

\newcommand{\beginsupplement}{%
        \setcounter{table}{0}
        \renewcommand{\thetable}{S\arabic{table}}%
        \setcounter{figure}{0}
        \renewcommand{\thefigure}{S\arabic{figure}}%
        \renewcommand{\theequation}{S\arabic{equation}}
        \setcounter{section}{0}
        \renewcommand{\thesection}{S\arabic{section}}
     }

\begin{document}
%
\title{3D Scene Creation and Rendering via Rough Meshes: A Lighting Transfer Avenue}
%
%
%
%

\author{Bowen Cai$^{\star, 1}$, Yujie Li$^{\star, 1}$, Yuqin Liang$^{1}$, Rongfei Jia$^{1}$, Binqiang Zhao$^{1}$, Mingming Gong$^{2}$, Huan Fu$^{1}$
\IEEEcompsocitemizethanks{\IEEEcompsocthanksitem Bowen Cai and Yujie Li contribute equally to this paper.\protect\\
\IEEEcompsocthanksitem 1. Tao Technology Department, Alibaba Group. \\
\IEEEcompsocthanksitem 2. School of Mathematics and Statistics, The University of Melbourne.}
\thanks{Manuscript received April 19, 2005; revised August 26, 2015.}}

%
%

\markboth{Journal of \LaTeX\ Class Files,~Vol.~14, No.~8, August~2015}%
{Shell \MakeLowercase{\textit{et al.}}: Bare Demo of IEEEtran.cls for Computer Society Journals}
%



\IEEEtitleabstractindextext{%
\begin{abstract}
This paper studies how to flexibly integrate reconstructed 3D models into practical 3D modeling pipelines such as 3D scene creation and rendering. Due to the technical difficulty, one can only obtain rough 3D models (R3DMs) for most real objects using existing 3D reconstruction techniques. As a result, physically-based rendering (PBR) would render low-quality images or videos for scenes that are constructed by R3DMs. One promising solution would be representing real-world objects as Neural Fields such as NeRFs, which are able to generate photo-realistic renderings of an object under desired viewpoints. However, a drawback is that the synthesized views through Neural Fields Rendering (NFR) cannot reflect the simulated lighting details on R3DMs in PBR pipelines, especially when object interactions in the 3D scene creation cause local shadows. To solve this dilemma, we propose a lighting transfer network (LighTNet) to bridge NFR and PBR, such that they can benefit from each other. LighTNet reasons about a simplified image composition model, remedies the uneven surface issue caused by R3DMs, and is empowered by several perceptual-motivated constraints and a new Lab angle loss which enhances the contrast between lighting strength and colors. Comparisons demonstrate that LighTNet is superior in synthesizing impressive lighting, and is promising in pushing NFR further in practical 3D modeling workflows.
\end{abstract}

}

\maketitle

\IEEEdisplaynontitleabstractindextext

%
\IEEEpeerreviewmaketitle

\IEEEraisesectionheading{\section{Introduction}\label{sec:introduction}}

%
%
%
%
\IEEEPARstart{T}{he} computer vision and graphics communities have put tremendous efforts into studying objects' representation methods for 3D modeling over the past years. In practical 3D modeling pipelines such as 3D scene designing, augmented reality (AR), and robotics, objects are usually represented as 3D CAD meshes combined with their materials and texture atlases (denoted as 3DMs). However, even the state-of-the-art (SOTA) 3D reconstruction methods do not have very accurate mesh reconstructions \cite{xu2019disn,mescheder2019occupancy,huang2018deepmvs,niemeyer2020differentiable,yariv2020multiview}. As a consequence, physical-based rendering (PBR) can only render low-quality content from these rough 3D models (R3DMs).

\begin{figure*}[th!]
    \centering
    \includegraphics[width=1.0\textwidth]{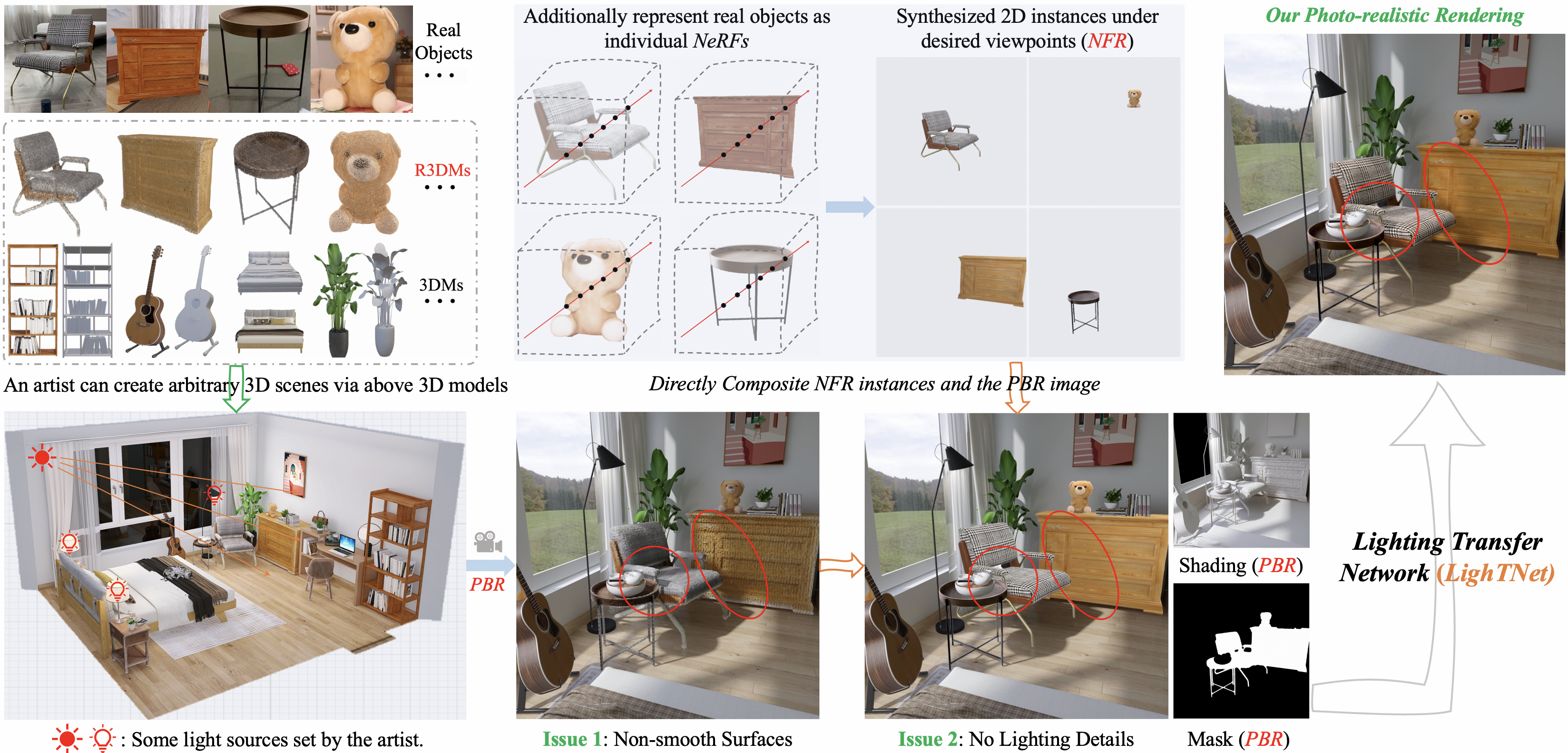}
    \caption{{\textbf{A Lighting Transfer Avenue.} \emph{Left (Problem):} Given some reconstructed rough 3D models (R3DMs) and designed 3D CAD models (3DMs), artists can use them to \textcolor{black}{create any 3D scenes and freely perform arbitrary lighting simulation} for each created scene. A physically-based rendering (PBR) system can only render low-quality images or videos for these scenes. Our goal is to render high-quality content from these possible scenes \textcolor{black}{without training (or fitting) the newly created 3D scenes}. \emph{Right (Solution):} As an example, if we have pre-obtained a neural fields representation (\emph{e.g.} NeRF \cite{mildenhall2020nerf}) for each real object, we can synthesize object instances for R3DMs in impressive quality through neural fields rendering (NFR). Unluckily, NFR instances cannot reflect the simulated lighting details (\emph{e.g.}, local shadows) on R3DMs. We propose a lighting transfer network (LighTNet) to bridge NFR and PBR, such that they can benefit from each other. In practice, LighTNet is trained once in a dataset and can be used for all the newly created 3D scenes with both seen and unseen R3DMs and arbitrary lighting (See ``Generalizing to Real-Lighting" in Fig.~\ref{fig:real}).}}
    \label{fig:relighting}
\end{figure*}

In this paper, we study how to flexibly integrate reconstructed 3D models into practical 3D modeling pipelines such as 3D scene creation and rendering. More specifically, we consider a practical setting in which one can utilize R3DMs (or R3DMs together with 3DMs) to create any scenes and assign arbitrary lighting to each created scene. Our goal is to render high-quality content from these possible scenes without training (or fitting) the newly created scenes. A possible solution is to represent real-world objects as \emph{Neural Fields} such as NeRF \cite{mildenhall2020nerf} in addition to R3DMs. \HF{Specifically, in the design process, the primary role of a 3D model lies in interaction (with the environment and with light). If we can obtain information about these interactions (such as the effects of various lights on objects) through physical rendering, and then combine this with NeRF to acquire high-precision appearance information, we can use neural networks to simulate the process of rendering element composition. This enables us to mitigate the impact of rough 3D models and obtain high-precision scene renderings.} As shown in Fig.~\ref{fig:relighting}, given both the explicit representations (R3DMs) and implicit representations (NeRFs) of several real-world objects, artists can create unlimited 3D scenes in graphics software, then freely render high-quality images and videos by simply compositing PBR images and NFR images.

In further, artists may perform free lighting simulation to their created 3D scenes (\emph{e.g.} setting several strong light sources) to capture realistic renderings. The remained question is that the above rendering routing cannot reflect the simulated lighting details on R3DMs in PBR pipelines, especially when object interactions in the 3D scene creation cause local shadows. Some works on neural light fields \cite{oechsle2020learning,zhang2021neural,bi2020neural,thies2019deferred,physg2020,nerfactor} learn to simulate arbitrary lighting to an object or a scene. Still, they cannot model the complex local shadows without fitting a static scene under many possible lighting conditions. Thus, these light field methods would require training on each newly created scene under different lighting conditions, which is impractical for real-world applications. 

 To solve the dilemma, we propose a Lighting Transfer Network (LighTNet) to bridge NFR and PBR, such that they can benefit from each other. LighTNet takes ``Shading" rendered from a PBR system and a synthesized image by NFR techniques (\emph{e.g.} NeRF) as input and outputs photo-realistic renderings with rich lighting details. Taking inspiration from the image composition process in V-Ray \cite{vraydoc}, we prudently reformulate it to remedy the non-smooth ``Shading" surfaces caused by R3DMs as well as better preserve lighting details. Furthermore, we propose perceptual-motivated constraints to optimize LightNet and introduce a novel \emph{Lab} Angle loss which can enhance the contrast between lighting strength and colors. To train LighTNet, we synthesize R3DMs by injecting random noises into 3DMs and use these $<$R3DM, 3DM$>$ pairs as the training data. Once LighTNet is learned on the training data, it can be used for arbitrary newly created 3D scenes with both seen and unseen R3DMs. Experiments show that LighTNet is superior in synthesizing impressive lighting details.

In summary, our main contributions are as follows:
\begin{itemize}
  \item We present a lighting transfer avenue that allows artists to create arbitrary 3D scenes, flexibly simulate lighting, and freely render photo-realistic images and videos via R3DMs and 3DMs in any graphic software.
  \item \HF{With the pathway, we develop a lighting transfer network (LighTNet) leveraging a prudently reformulated image composition formulation. This network effectively bridges the lighting gap between PBR and NFR, showing promise in addressing the non-smooth "Shading" surfaces resulting from R3DMs.}
  \item We introduce a \emph{Lab} Angle loss to enhance the contrast between lighting strength and colors which can further improve the rendering quality.
\end{itemize}

\section{Related Work}
\subsection{3D Object Reconstruction}

Typical SFM and MVS approches \cite{andrew2001multiple,schoenberger2016sfm,schoenberger2016mvs} can reconstruct 3D meshes of objects that are with rich textures in reasonable quality. Leveraging large database, researchers exploit deep neural networks to reconstruct point clouds \cite{achlioptas2018learning,fan2017point,thomas2019kpconv,yang2019pointflow}, voxel grids \cite{brock2016generative,choy20163d,gadelha20173d,xie2019pix2vox,wu2016learning}, and meshes \cite{kanazawa2018learning,liao2018deep,pan2019deep,wang2018pixel2mesh} from single or multiple images. Other works show learning implicit representations for objects is a promising avenue \cite{atzmon2019controlling,xu2019disn,saito2019pifu,park2019deepsdf,michalkiewicz2019implicit,mescheder2019occupancy,wang2021neus,oechsle2021unisurf,yariv2021volume}. For example, IDR \cite{yariv2020multiview} and DVR \cite{niemeyer2020differentiable} take advantage of differentiable rendering formulation for implicit shape and texture representations and show the possibility of recovering smooth surfaces for objects with rich textures from a set of posed images. They cannot handle many real-world cases, such as big items with flattened areas (e.g. furniture). Besides, they fall into the neural rendering category, thus would also benefit from the lighting transfer avenue. 

To the best of our knowledge, no high-performing solution can automatically reconstruct perfect meshes and their UV texture atlases for real-world objects. Moreover, even if we can obtain an ideal 3D model with a perfect topology, we also need to rebuild its UV textures and materials. Unfortunately, texture and material recovery are currently receiving relatively poor attention, and the progress is not smooth.

\begin{figure*}[t!]
    \centering
    \includegraphics[width=1.0\textwidth]{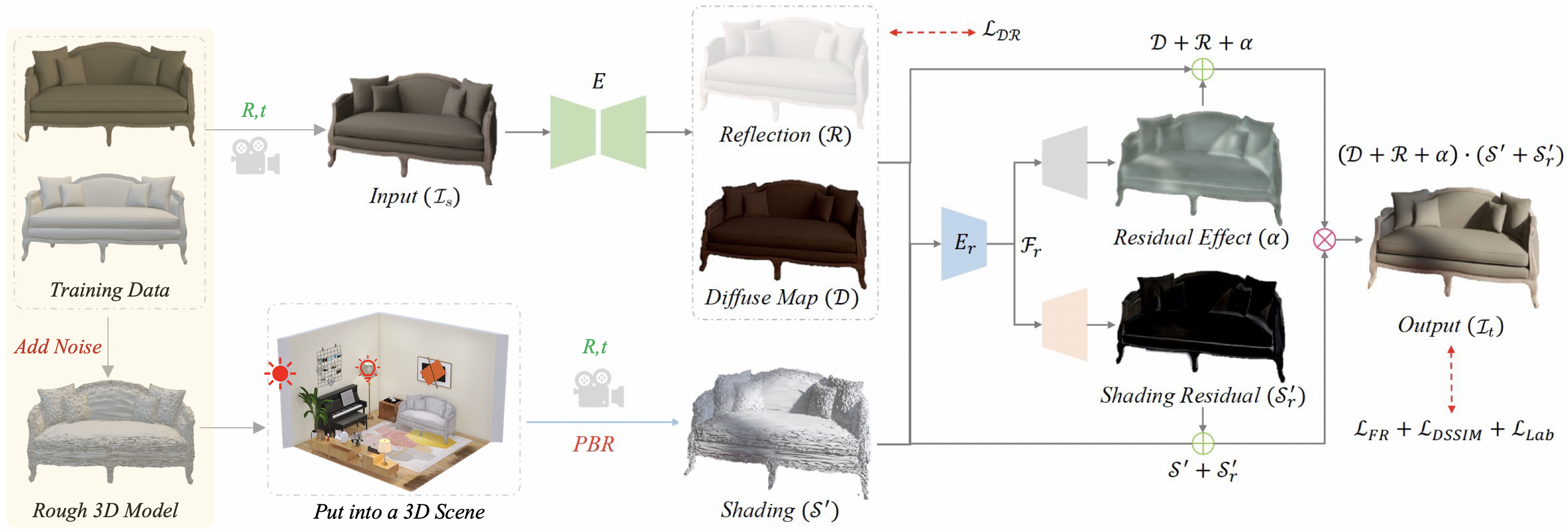}
    \caption{{\textbf{Training a LighTNet.} LighTNet aims to transfer the lighting details from an imperfect shading map $\mathcal{S}'$ to the corresponding image $\mathcal{I}_s$. It reasons about the reformulated image composition model $\mathcal{I}_t = (\mathcal{D} + \mathcal{R} + \alpha) \cdot (\mathcal{S}' + \mathcal{S}'_r)$. The yellow (left) part shows the $<$R3DM, 3DM$>$ pairs generation process, and is only included in the training process. Once optimized, LighTNet can be used for any newly created 3D scenes with both seen and unseen R3DMs and support free lighting simulation. In the inference phase, $\mathcal{I}_s$ of an object is the 2D instance synthesized by a trained NeRF or any other high-performing free view synthesis formulations (See Fig.~\ref{fig:relighting} and Fig.~\ref{fig:nvr3d-sd}).
    }}
    \label{fig:lightnet}
\end{figure*}

\begin{figure}[t!]
    \centering
    \includegraphics[width=0.48\textwidth]{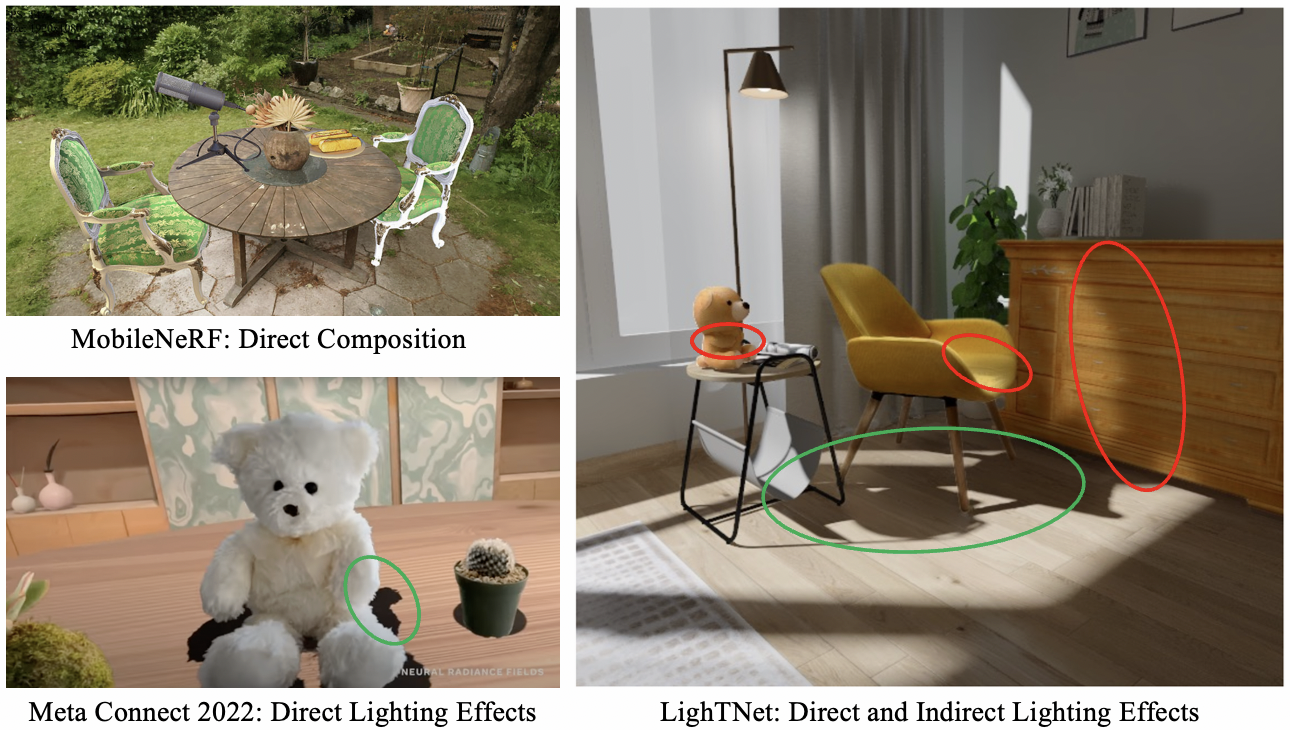}
    \caption{{\textbf{Compositing Individual NeRF Objects.} Several concurrent works \cite{chen2022mobilenerf,metaconnect22} show it's possible to represent real-world objects as individual NeRFs and R3DMs for freely 3D scene creation and rendering. LighTNet goes a futher step by considering the indirect lighting effects such as local shadows on R3DMs caused by objects interactions.
    }}
    \label{fig:scene-comp}
\end{figure}

\subsection{Neural Rendering Leveraging NeRFs} 
Recent advances show neural fields representations are promising to describe scenes, and support rendering photo-realistic images of the fitted scenes under desired viewpoints \cite{mildenhall2020nerf,chen2022mobilenerf,martin2021nerf,neff2021donerf,niemeyer2020giraffe,peng2021neural,pumarola2020d,zhang2020nerf++,Suhail_2022_CVPR,Barron_2021_ICCV,yu2021plenoctrees,Sun_2022_CVPR,Chen2022ECCV,Hu_2022_CVPR,Barron_2022_CVPR,Verbin_2022_CVPR}. Concurrently, MobileNeRF \cite{chen2022mobilenerf} has performed scene edition application by representing real-world objects as individual NeRFs and R3DMs. Meta in Meta Connect 2022 \cite{metaconnect22} demonstrates this routine supports direct shadow simulations. The proposed lighting transfer avenue goes a further step by modeling the indirect lighting effects, such as local shadows on R3DMs caused by object-to-object interactions. 

There are several works \cite{philip2021free,ye2022intrinsicnerf,tancik2022block} that have also exploited free scene lighting editing. Unlike these approaches that would first perform per-scene optimization before editing and rendering, LighTNet is a generic solution that can be directly integrated into practical 3D modeling pipelines for scene creation and rendering without per-scene optimization. It's worth mentioning that some works study inverse rendering with implicit neural representations that enable material editing and free view relighting of their reconstructed scenes (or objects) \cite{physg2020,zhang2021nerfactor,boss2020nerd,srinivasan2020nerv,zhang2022modeling}. Our setting is totally different from theirs. For example, they can only synthesize local shadows caused by \emph{self-occlusion} of its optimized single scene (or object). In fact, they have not considered compositing individual NeRFs to freely create and edit 3D scenes, thus have not handle the possible indirect lighting effects caused by objects interaction. We refer to the ``Discussion" section for more explanation about the differences.








\section{Lighting Transfer Network (LighTNet)}
\label{sec:lightnet}
As shown in Fig.~\ref{fig:lightnet}, the goal of LighTNet is to transfer the lighting details from an imperfect shading map $\mathcal{S}'$ to the corresponding image $\mathcal{I}_s$. We will start with a brief introduction to the simplified image composition formulation in Sec.~\ref{subsec:rid}, which is the theoretical basis of LighTNet. Then, we explain the network architecture in Sec.~\ref{subsec:arch}. Finally, in Sec.~\ref{subsec:obj}, we introduce the proposed \emph{Lab Angle Loss} and other involved objectives. 

Importantly, in Sec.~\ref{subsec:nvr3d-lightnet}, we will take a specific example to explain the \emph{Lighting Transfer} avenue. We show how a trained LighTNet allows us to flexibly create 3D scenes, edit lighting, and render high-quality images and videos via R3DMs and 3DMs.

\begin{figure*}[th!]
    \centering
    \includegraphics[width=1.0\textwidth]{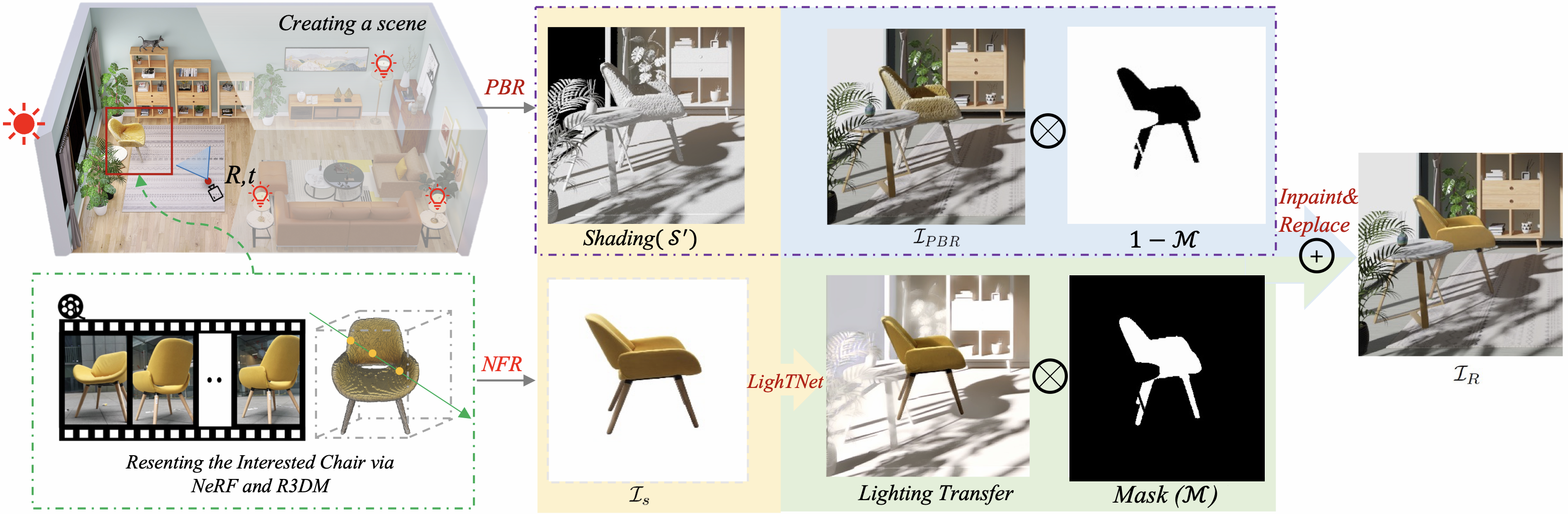}
    \caption{{\textbf{3D Scene Creation and Rendering via R3DMs.} We can represent real-world objects as individual NeRFs and R3DMs, and freely composite them to create unlimited 3D scenes. After lighting editing by artists, LighTNet can transfer direct and indirect lighting effects on R3DMs (e.g. $\mathcal{S}'$) to the corresponding NFR instances (e.g. $\mathcal{I}_s$). See Sec.~\ref{subsec:nvr3d-lightnet} for the detailed explanation.
    }}
    \label{fig:nvr3d-sd}
\end{figure*}

\subsection{Preliminaries: Image Composition}
\label{subsec:rid}

An image can be expressed as the point-wise product between its shading $\mathcal{S}$ and albedo $\mathcal{A}$, \emph{i.e.,} $\mathcal{I} = \mathcal{A} \cdot \mathcal{S}$, as discussed in \cite{janner2017self,ma2018single,zhou2019glosh,liu2020unsupervised}. $\mathcal{A}$ is often simplified as a diffuse map which shows the base colors and textures used in materials with no lighting information. However, the render equation \cite{kajiya1986rendering} in general physically-based renders tells us $\mathcal{A}$ should encode other material properties such as refraction and specularity. 
We follow the compositing process and definitions in V-Ray \cite{vraydoc} and simplify its formulation as 
\begin{equation}
\begin{split}
  \mathcal{I} = \mathcal{D} \cdot \mathcal{S}  + \mathcal{R} \cdot \mathcal{R}_l + \alpha_2,
\end{split}
\end{equation}
where $\mathcal{D}$ is the diffuse map, $\mathcal{S}$ is all the raw lighting (both direct and indirect ) in the scene and we regard it as ``Shading" in this paper, $\mathcal{R}$ defines the strength of the reflection of the materials, $\mathcal{R}_l$ stores reflection information calculated from the materials' reflection values in the scene, and $\alpha_2$ provides the interactive effects between other material properties and lighting. As $\mathcal{S}$ encodes the per-pixel lighting of a scene, we approximate $\mathcal{R}_l$ via $\mathcal{S} + \alpha_1$. Going a further step, we find that rendering the supervision information $\mathcal{S}+\alpha_1$ and $\alpha_2$ is impractical because we only have a rough geometry R3DM. Besides, learning them separately would increase our framework complexity. We thus further simplify the formulation as
\begin{equation}\label{eqn:rid}
\begin{split}
  \mathcal{I} = (\mathcal{D} + \mathcal{R} + \alpha) \cdot \mathcal{S}
\end{split}
\end{equation}
by regarding ${(R \cdot \alpha_1 + \alpha_2)} / {S}$ as a packed residual effect $\alpha$.


\subsection{Architecture}
\label{subsec:arch}
Our lighting transfer network (LighTNet) is developed based on the formulation $\mathcal{I} = (\mathcal{D} + \mathcal{R} + \alpha) \cdot \mathcal{S}$. Given a training sample $(\mathcal{I}_s, \mathcal{S}', \bar{\mathcal{I}}_{t}, \bar{\mathcal{D}}, \bar{\mathcal{R}})$, where $\bar{\mathcal{I}}_{t}$, $\bar{\mathcal{D}}$, and $\bar{\mathcal{R}}$ are the ground-truth images, LighTNet takes $\mathcal{I}_s$ and $\mathcal{S}'$ as inputs, and target at reconstructing $\bar{\mathcal{I}}_{t}$. See Sec.~\ref{subsec:data-const} for details about the training samples capturing process. 

We utilize an encoder-decoder network $E$ to estimate both the diffuse map $\mathcal{D}$ and the reflection strength $\mathcal{R}$ from $\mathcal{I}_s$. We learn $\mathcal{R}$ and $\mathcal{D}$ in a supervised manner using:
\begin{equation}
\begin{split}
  \mathcal{L}_{\mathcal{D}\mathcal{R}} = | \mathcal{R} - \bar{\mathcal{R}} | + | \mathcal{D} - \bar{\mathcal{D}} |.
\end{split}
\end{equation}

After that, the remained major issue is that the shading $S'$ is not smooth since its R3DM's surfaces are uneven. We know that shading is determined by surface normal and illumination \cite{zhou2019glosh}. We clarify $\mathcal{D}$ and $\mathcal{R}$ mapping from $\mathcal{I}_s$ usually imply smooth normal information that could remedy $\mathcal{S}'$. Towards the purpose, we obtain an intermediate representation by concatenating $\mathcal{S}'$, $\mathcal{D}$, and $\mathcal{R}$ together, and take an encoder network ($E_r$) to map it to a feature $\mathcal{F}_{r}$. With $\mathcal{F}_{r}$, a straightforward option is to directly predict a smooth shading $\mathcal{S}$. In our experiments, we find a smoother $\mathcal{S}$ could be captured following $\mathcal{S} = \mathcal{S}' + \mathcal{S}'_r$, where $\mathcal{S}'_r$ is the learned residual from $\mathcal{F}_{r}$ via a decoder network ($D_{\mathcal{S}'}$).

Finally, as analyzed before, the model $ \mathcal{I} = (\mathcal{D} + \mathcal{R}) \cdot \mathcal{S}$ cannot describe the full lighting effects. Especially, our experiments in Fig.~\ref{fig:decomp} show it cannot well preserve shadows. We thus follow Eqn.~\ref{eqn:rid} to directly predict a residual effect $\alpha$ from $\mathcal{F}_{r}$ through another decoder network ($D_{\alpha}$). Considering all above, the target image can be composited as
\begin{equation}\label{eqn:lt}
\begin{split}
  \mathcal{I}_t = (\mathcal{D} + \mathcal{R} + \alpha) \cdot (\mathcal{S}' + \mathcal{S}'_r).
\end{split}
\end{equation}

\begin{figure*}[th!]
    \centering
    \includegraphics[width=0.9\textwidth]{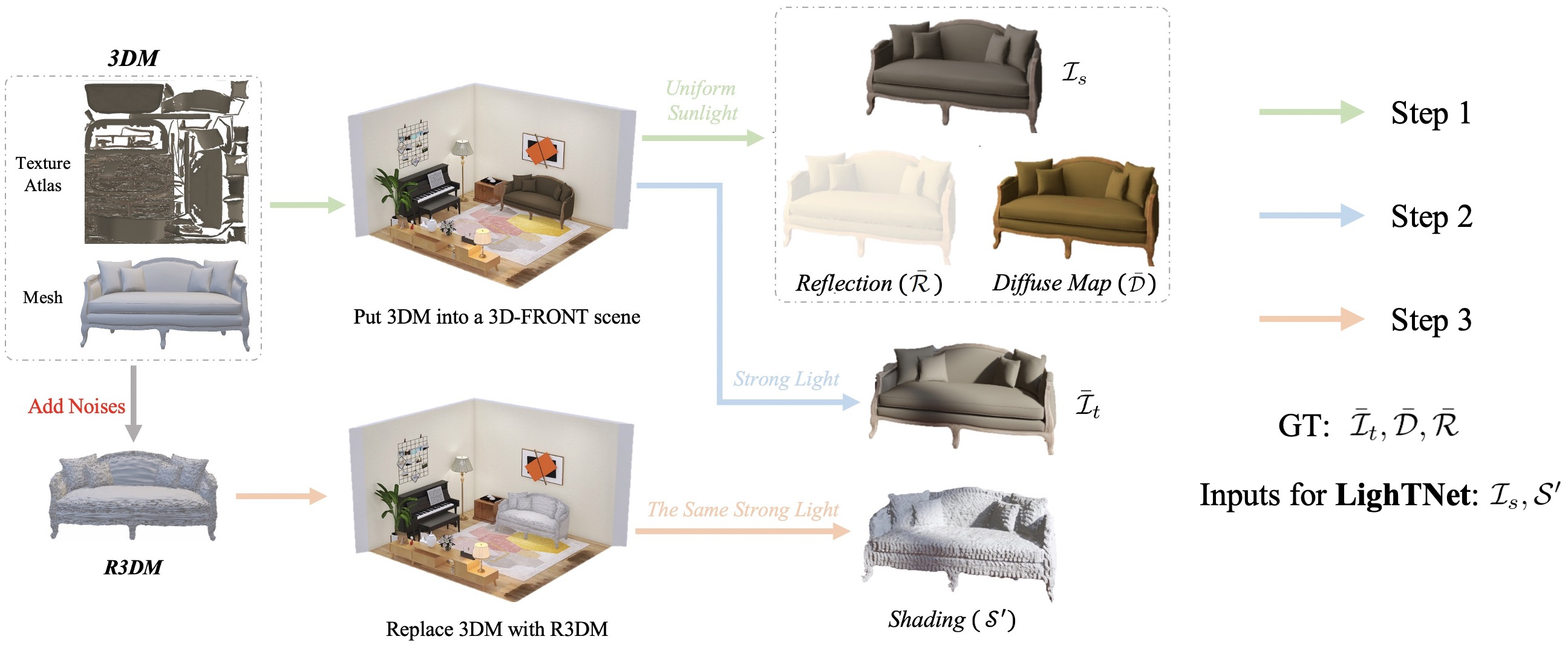}
    \caption{\textbf{Training Set Construction.} {We take the ``Sofa" case as an example to show how to capture a training sample $\{ (\mathcal{I}_s, \mathcal{S}', \bar{\mathcal{I}}_{t}, \bar{\mathcal{D}}, \bar{\mathcal{R}}\}$ via 3D CAD models and 3D scenes. The elements are rendered by Blender\cite{blender}. LighTNet is trained once on the 3DF-Lighting training set, and can be used for all the newly created scenes with both seen and unseen R3DMs and arbitrary lighting.}}
    \label{subfig:traindata}
\end{figure*}

\subsection{Objectives}
\label{subsec:obj}

In previous relighting efforts, the \emph{L1} photometric loss $\mathcal{L}_{\emph{L1}}$ ($|\mathcal{I}_t - \bar{\mathcal{I}}_t|$ or $|\log\mathcal{I}_t - \log\bar{\mathcal{I}}_t|$) was commonly used as a major term to preserve the basic image content in the reconstruction process \cite{sengupta2018sfsnet,yu2019inverserendernet,li2018learning}. However, we find in our experiments it will degrade the lighting transfer ability of LighTNet as shown in Fig.~\ref{fig:objective}. A possible reason is that it pushes the learning procedure to focus more on reducing the color differences instead of local lighting discrepancies. We thus propose to minimize the following losses that are closely related to perceptual quality and lighting effects. 
\newline

\noindent \textbf{Feature Reconstruction Loss.} The feature reconstitution loss \cite{johnson2016perceptual} encourages $\mathcal{I}_t$ to be perceptually similar to $\bar{\mathcal{I}}_t$ by matching their semantic features.  We take VGG-19 \cite{simonyan2014very} pretrained on ImageNet \cite{deng2009imagenet} as the feature extractor $\phi$ and denote $\phi_j(x)$ as the output of the $jth$ convolution block. The feature reconstruction loss is expressed as:
\begin{equation}
\begin{split}
  \mathcal{L}_{FR} = \frac{1}{C_j*H_j*W_j} \sum_{c, h, w} \| \phi_j(\mathcal{I}_t) - \phi_j(\bar{\mathcal{I}}_t) \|,
\end{split}
\end{equation}
where $C_j*H_j*W_j$ is the feature dimensions of $\phi_j(x)$. In this paper, we utilize activations of the third convolution block ($j=3$) to compute $\mathcal{L}_{FR}$.
\newline

\noindent \textbf{Structural Dissimilarity.} SSIM \cite{wang2004image} is another perceptual-motivated metric that measures structural similarity between two images. We take the structural dissimilarity (DSSIM) as a measure following the success in \cite{nestmeyer2020learning}:
\begin{equation}
\begin{split}
  \mathcal{L}_{DSSIM} = \frac{1 - \text{SSIM}(\mathcal{I}_t, \bar{\mathcal{I}}_t)}{2},
\end{split}
\end{equation}

\noindent \textbf{\emph{Lab} Angle Loss.} We find that LighTNet optimized with aforementioned loss terms would produce images with darker global brightness as shown in Fig.~\ref{fig:objective}. A possible reason is that $\mathcal{L}_{FR}$ and $\mathcal{L}_{DSSIM}$ only enhance local perceptual quality while overlooking the lighting contrast. In the paper, we thus propose a novel \emph{Lab} Angle loss to consider the pixel-wise ratio between lighting strength and colors as:
\begin{equation}
\begin{split}
  \mathcal{L}_{Lab} = \frac{1}{H*W}\sum_{h, w}\arccos{(\frac{\langle \varphi(\mathcal{I}_t)_{(h, w)} , \varphi(\bar{\mathcal{I}}_t)_{(h, w)} \rangle}{\| \langle \varphi(\mathcal{I}_t)_{(h, w)} , \varphi(\bar{\mathcal{I}}_t)_{(h, w)} \rangle \|})},
\end{split}
\end{equation}
where $\langle x, y \rangle$ denotes the inner product of vector $x$ and $y$,  $\varphi(\cdot)$ represents the \emph{RGB} to \emph{Lab} converter, $(h, w)$ is the spatial location, and $H*W$ is the image size.
\newline

\noindent \textbf{Full Objective.} Our LighTNet is optimized in an end-to-end fashion with the objective:
\begin{equation}
\begin{split}
  \mathcal{L} = \mathcal{L}_{\mathcal{D}\mathcal{R}} + \lambda_1\mathcal{L}_{FR} + \lambda_2\mathcal{L}_{DSSIM} + \lambda_3\mathcal{L}_{Lab},
\end{split}
\end{equation}
where the loss weights $\lambda_1$, $\lambda_2$, and $\lambda_3$, in all the experiments, are set to 0.05, 0.5, and 0.5, respectively.

\section{Rendering with LighTNet and R3DMs}
\label{subsec:nvr3d-lightnet}
In this section, we show how a trained LighTNet and R3DMs can be flexibly integrated into practical 3D modeling workflows such as 3D scene creation and rendering. For example, we are interested in a real yellow chair, as shown in Fig.~\ref{fig:nvr3d-sd}. Given its reconstructed R3DM and neural fields representation (NeRF \cite{mildenhall2020nerf} in this paper), we can create a 3D scene by putting the yellow chair's R3DM and some 3DMs into a 3D room. Here, both the room and the involved 3DMs have not been seen before. To showcase the scene, we would like to render a high-quality image, in which the involved 3D models are with rich lighting details. Towards the goal, we can set a high-energy light source, and render a scene image $\mathcal{I}_{PBR}$, a shading $\mathcal{S}'$, and an object's mask $\mathcal{M}$, under a good viewpoint. Simultaneously, we synthesize an image $\mathcal{I}_{s}$ with the same camera pose via NeRF. Then, we are able to capture a target image $\mathcal{I}_{t}$ by transferring lighting from $\mathcal{S}' \cdot \mathcal{M}$ to $\mathcal{I}_{s} \cdot \mathcal{M}$ via the trained LighTNet model. Finally, we replace $\mathcal{I}_{PBR} \cdot \mathcal{M}$ with $\mathcal{I}_{t} \cdot \mathcal{M}$ to obtain the final photo-realistic rendering $\mathcal{I}_{R}$. The complete process can be formulated as:
\begin{equation}
\begin{split}
  \mathcal{I}_{R} =\mathcal{I}_{PBR}\otimes(1 - \mathcal{M}) +  \mathcal{M}\otimes\text{LighTNet}(\mathcal{I}_s, \mathcal{S}'),
\end{split}
\end{equation}
where $\otimes$ is the element-wise production operation. In practice, there will be some regions in boundary areas that cannot be covered by $\mathcal{I}_{t}$. We directly fill these regions via a SOTA image inpainting technique \cite{nazeri2019edgeconnect}. Some qualitative results are shown in Fig.~\ref{fig:real}.
\begin{figure*}[t!]
    \centering
    \includegraphics[width=1.0\textwidth]{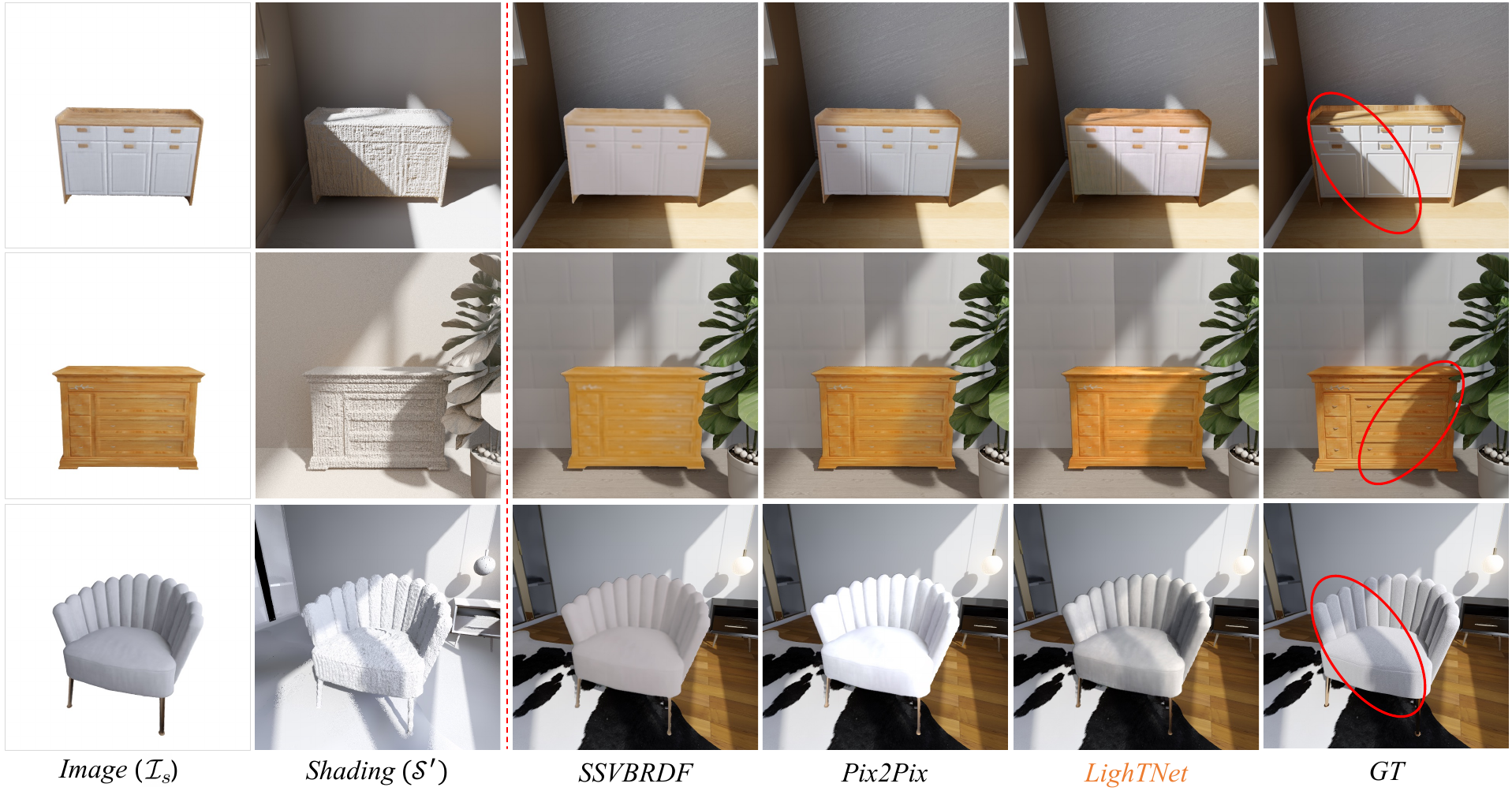}
    \caption{\textbf{Qualitative Comparisons on 3DF-Lighting.} {We make qualitative comparisons with the reformulated Pix2Pix \cite{isola2017image} and SSVBRDF \cite{li2018learning}. LighTNet achieves realistic relighting results with impressive shadow details. $\mathcal{I}_s$ and $\mathcal{S}'$ are rendered by NFR and PBR, respectively.
    }}
    \label{fig:comparisons}
\end{figure*}

\setlength\tabcolsep{4pt}
\begin{table}[t!]
\caption{\textbf{Quantitative Evaluation on 3DF-Lighting.} We use the proposed \emph{Lab} Angle distance and several widely studied metrics, including \emph{L1}-Norm, PSNR, and SSIM\cite{wang2004image}, to measure a method's lighting transfer ability. \HF{LighTNet and LighTNet$^\dag$ denote training LighTNet with Blender and NeRF images, respectively (Refer to Sec. 5.1.1).}}
\centering
\small
\begin{tabular}{ c  c  c  c  c }
\toprule
Method & \emph{L1}-Norm $\downarrow$ & PSNR $\uparrow$ & SSIM $\uparrow$ & \emph{Lab} Angle $\downarrow$ \\
\cmidrule(lr){1-1}\cmidrule(lr){2-5}
Pix2Pix \cite{isola2017image} & 0.0345 & 26.65 & 0.9042 & 0.4314 \\
DPR\cite{zhou2019deep} & 0.0399 & 25.39 & 0.8692 & 0.4576 \\
SSVBRDF \cite{li2018learning} & 0.0373 & 26.02 & 0.9040 & 0.3796 \\
\midrule
LighTNet & \textbf{0.0219} & \textbf{30.17} & 0.9142 & \textbf{0.3137} \\
\HF{LighTNet$^\dag$} & 0.0281 & 29.93 & \textbf{0.9203} & 0.3173 \\
\bottomrule
\end{tabular}
\label{tab:quant-eval}
\end{table}


\section{Experiments}
\label{sec:exp}

In this section, we conduct many experiments to examine the lighting transfer avenue. We first present the training and evaluation sets building processes in Sec.~\ref{subsec:data-const}. Then, we build several baselines and make benchmark comparisons with them in Sec.~\ref{subsec:comparison}. Finally, we perform various ablation studies to discuss our method in Sec.~\ref{subsec:ablation}.

\subsection{Datasets}
\label{subsec:data-const}

\subsubsection{3DF-Lighting} 
\label{subsubsec:3df-lighting}
\noindent \textbf{Training Set:} We take 50 3D scenes and the involved 30 3D CAD models (denoted as 3DMs) in 3D-FRONT \cite{fu20213d} to construct the training set. First, we need to recover these objects' rough 3D meshes (R3DMs). We simply adopt the mesh subdivision algorithm \cite{Loop_1987_7735,Zhou2018Open3d} to densify the CAD models' surfaces, then add random noise to each vertex. Second, for a specific object in a scene, we simulate uniform sunlight to the scene, randomly choose a viewpoint and render the object's diffuse map $\bar{\mathcal{D}}$, reflection strength $\bar{\mathcal{R}}$, and color image $\mathcal{I}_s$. Third, we randomly change the light source's position and increase the lighting energy to capture a target image $\bar{\mathcal{I}}_{t}$. Finally, we render the rough shading $S'$ by replacing the object's 3DM as its R3DM. Following the pipeline, we can construct a training set $\{ (\mathcal{I}_s, \mathcal{S}', \bar{\mathcal{I}}_{t}, \bar{\mathcal{D}}, \bar{\mathcal{R}}\}$. This paper takes Blender \cite{blender} with V-Ray plug-in as the render engine to secure these elements. \HF{We have also rendered $\mathcal{I}_s$ to maintain consistency between training and inference. As reported in Tab. 1, the "NeRF" setting (LighTNet$^\dag$) yields similar scores compared to the "Blender" setting (LighTNet).}
\newline

\noindent \textbf{Evaluation Set:} We build a test set using another 10 furniture shapes and 20 3D scenes from 3D-FRONT. We take one object as an example to present the test set building process. We randomly render 200 images from viewpoints sampled on a full sphere to learn its NeRF and R3DM. For each 3D scene, we put the object's R3DM into the scene and randomly render thirty $\mathcal{S}'$ and $\mathcal{M}$. Simultaneously, we synthesize the corresponding thirty $\mathcal{I}_s$ using its NeRF. Finally, we render the thirty $\bar{\mathcal{I}}_t$ (ground truth images) at size $800\times800$ by replacing the R3DM with its 3DM. Each 3D scene's light source and energy are pre-defined. Through the workflow, we construct a test set with 6,000 samples $\{(\mathcal{I}_s, \mathcal{S}', \bar{\mathcal{I}}_t, \mathcal{M})\}$. We pre-assign V-Ray materials to each 3D model (R3DM and R3D) manually. 

\begin{figure*}[t!]
    \centering
    \includegraphics[width=0.9\textwidth]{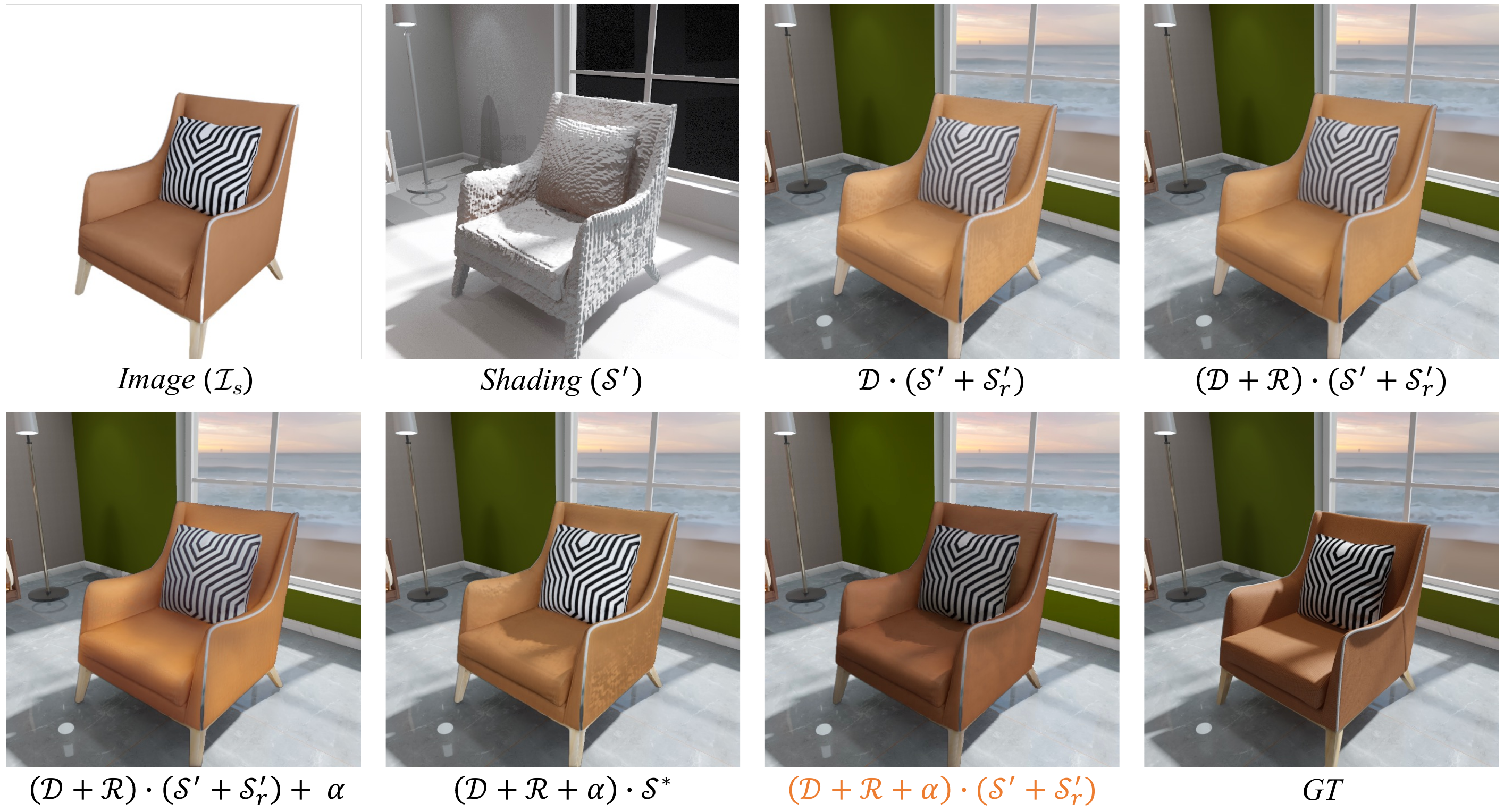}
    \caption{{We qualitatively evaluate the lighting transfer ability of the image composition variants. $(\mathcal{D} + \mathcal{R} + \alpha) \cdot \mathcal{S}$ would be a much better choice for simulating the PBR compositing process. 
    }}
    \label{fig:decomp}
\end{figure*}

\setlength\tabcolsep{8pt}
\begin{table*}[t!]
\caption{We find that optimizing LighTNet with a \emph{L1} photometric loss ($\mathcal{L}_{L1}$ = $|\mathcal{I}_t - \bar{\mathcal{I}}_t|$) would yield a degenerate performance.}
\centering
\small
\begin{tabular}{ c c c  c  c  c  c  c }
\toprule
\multicolumn{4}{c}{Objective} & \multicolumn{4}{c}{Metric} \\
\cmidrule(lr){1-4}\cmidrule(lr){5-8}
$\mathcal{L}_{FR}$ & $\mathcal{L}_{DSSIM}$ & $\mathcal{L}_{Lab}$ & $\mathcal{L}_{L1}$ & \emph{L1}-Norm $\downarrow$ & PSNR $\uparrow$ & SSIM $\uparrow$ & \emph{Lab} Angle $\downarrow$ \\
\midrule\midrule
 &  &  & $\surd$ & 0.0277 & 28.37 &0.8774 & 0.3551 \\
$\surd$ &  &  &  & 0.0254 & 29.06 & 0.9006 & 0.3717 \\
$\surd$ & $\surd$ &  &  & 0.0229 & 29.83 & 0.9129 & 0.3317 \\
$\surd$ & $\surd$ & $\surd$ &  & \textbf{0.0219} & \textbf{30.17} & \textbf{0.9142} & \textbf{0.3137} \\
$\surd$ & $\surd$ & $\surd$ & $\surd$ & 0.0240 & 29.37 & 0.9102 & 0.3426 \\
\bottomrule
\end{tabular}
\label{tab:objectives}
\end{table*}

\begin{figure}[th]
    \centering
    \includegraphics[width=0.48\textwidth]{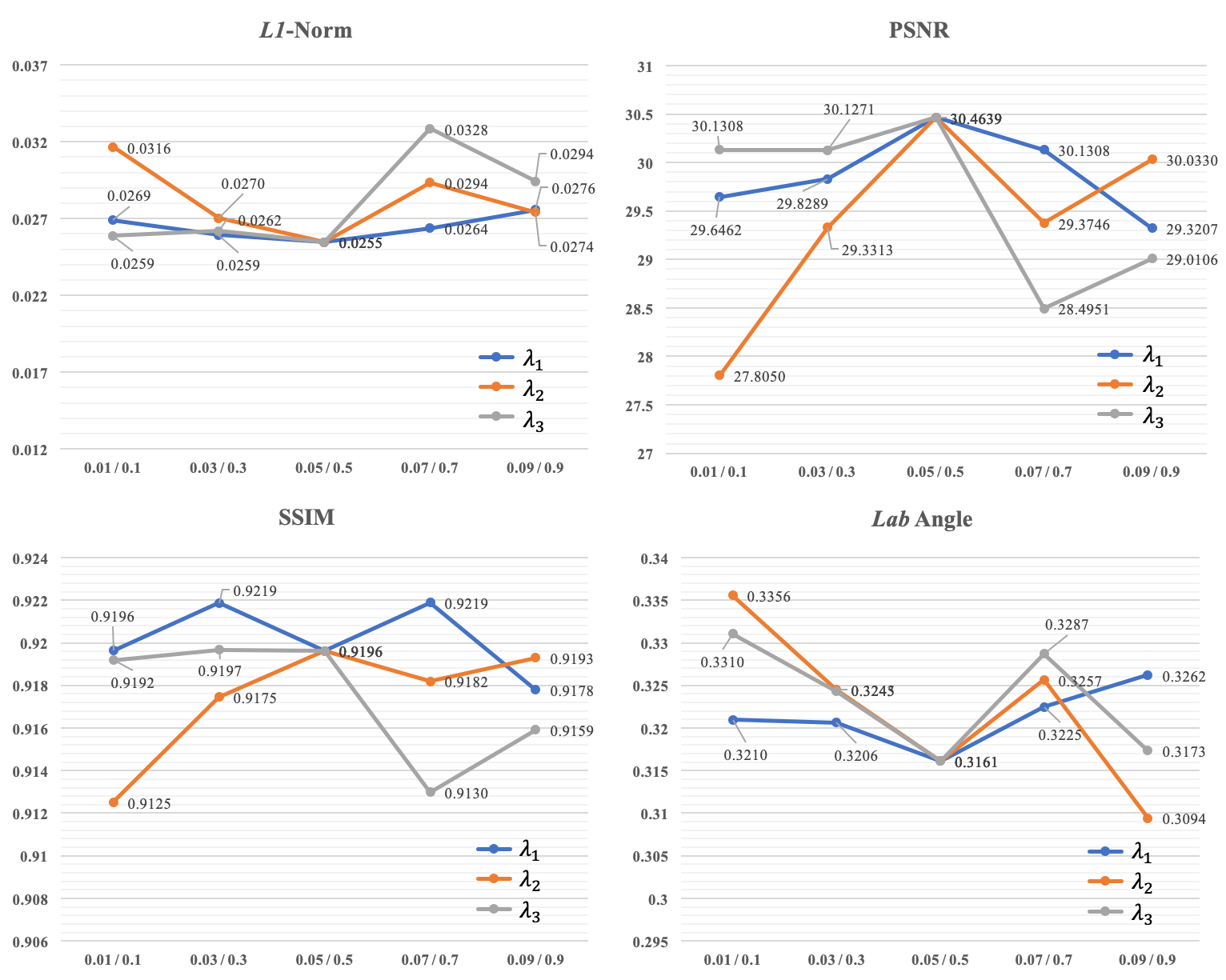}
    \caption{\HF{\textbf{Weights for loss terms.} We study  $\lambda_1\mathcal{L}_{FR} + \lambda_2\mathcal{L}_{DSSIM} + \lambda_3\mathcal{L}_{Lab}$ through a controlled variable technique. See Sec.~\ref{subsec:ablation} for the setting of this ablation study.}}
    \label{fig:loss_weights}
\end{figure}

\subsubsection{Generalizing to Real-Lighting} 
We also conduct qualitative evaluation on a real dataset named Real-Lighting. Specifically, we capture some object-centric videos via a mobile phone, and reconstruct these objects via NeRF. We create some 3D scenes using these objects' R3DMs and other 3DMs . Some rendered images of these scenes are presented in Fig.~\ref{fig:real}. We can see the lighting details have been successfully preserved. The LighTNet model is trained only on the 3DF-Lighting train set. All the scenes and objects in Real-Lighting have not been seen previously. 

\HF{Moreover, we choose two individuals from the ActorsHQ~\cite{isik2023humanrf} dataset and incoperate two real-world 360 scenes from Mip-NeRF 360~\cite{Barron_2022_CVPR} for our experiment. Using Instant-NGP~\cite{muller2022instant}, we reconstructed the two individuals, and for the two scenes, we employed Mip-NeRF 360~\cite{Barron_2022_CVPR}. Subsequently, we integrated the digital representations of the individuals and objects into the reconstructed 3D scenes, forming new composite 3D scenes. Light sources were randomly placed in these new 3D scenes. That means, all the content in the new scenes are reconstructed. The qualitative results, depicted in Fig.~\ref{fig:real_scene} and the accompanying video, highlight the adaptability of our method to both human subjects and reconstructed real-world 3D scenes.}

\subsection{Benchmark Comparisons}
\label{subsec:comparison}
\noindent \textbf{Building Baselines.} We build baselines by reformulating three works, including Pix2Pix \cite{isola2017image}, DPR \cite{zhou2019deep}, and SSVBRDF \cite{li2018learning}, to study the lighting transfer setting. \HF{Our research addresses a novel problem arising from the practical 3D house design and rendering workflow. With the proposed lighting transfer pathway depicted in Fig. 1, our objective is to transfer lighting details from the "Shading" map ($\mathcal{S}'$) to the NeRF rendering ($\mathcal{I}_s$) (refer to Fig. 4). While we accomplish this task through an image composition formulation based on the V-Ray render engine's compositing process, it can also be viewed as a conditional image-to-image translation problem. Hence, Pix2Pix \cite{isola2017image} is selected as one of our baselines. We learn the mapping from $\mathcal{I}_s\oplus\mathcal{S}'$ to $\mathcal{I}_t$, where $\oplus$ is the concatenate operation along the channel dimension. Considering DPR \cite{zhou2019deep} and SSVBRDF \cite{li2018learning}, they belong to the realm of inverse rendering, predicting BRDF from single images and supporting image relighting. To decompose BRDF, they require predicting lighting (usually environment lighting or spherical harmonics) for a given single image. Consequently, we can integrate them into the proposed lighting transfer avenue by directly rendering the scenes' spherical harmonics lighting through PBR and using it directly in the relighting process.} All the methods (including LighTNet) have been trained on the 3DF-Lighting train set.
\newline

\noindent \textbf{Performance.} To measure the lighting synthesis ability, we take \emph{L1}-Norm, PSNR, SSIM \cite{wang2004image}, and our \emph{Lab} Angle loss as the metrics. From the scores presented in Table~\ref{tab:quant-eval}, our LighTNet outperforms the compared methods by a large margin. Especially, while the best PSNR and \emph{L1}-Norm obtained by the baselines are 26.65 and 0.0345, LighTNet significantly improves them to 30.17 and 0.0219. It is not surprising since (1) DPR and SSVBRDF focus more on modeling global illumination, and (2) transferring lighting from shading with uneven surfaces is more challenging. Several qualitative comparisons are reported in Fig.~\ref{fig:comparisons}. LighTNet achieves realistic relighting results with impressive shadow details. In Fig.~\ref{fig:real}, we illustrate some further examples of our approach generalizing to real objects, using the LighTNet model only trained on 3DF-Lighting.

\subsection{Ablation Studies}
\label{subsec:ablation}
We argue that a slight numerical gain over the studied metrics may imply an improved visual experience since lighting is a detailed effect. We refer to the supplemental material for more qualitative comparisons.
\newline

\begin{figure*}[th!]
    \centering
    \includegraphics[width=1.0\textwidth]{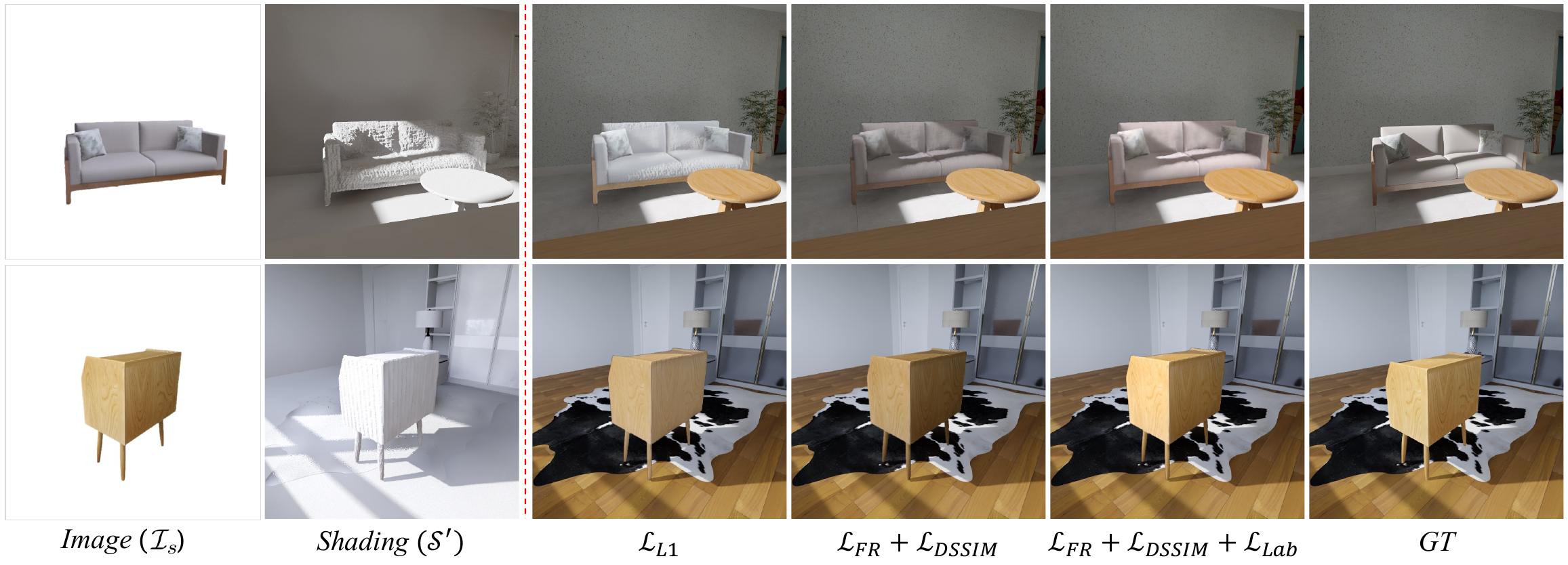}
    \caption{{We qualitatively discuss the objectives in Sec.~\ref{subsec:obj}. $\mathcal{L}_{FR} + \mathcal{L}_{DSSIM}$ yields a notable improvement in dealing with uneven shading surface and local shadows compared to $\mathcal{L}_{L1}$. $\mathcal{L}_{Lab}$ could further enhance the lighting effects. 
    }}
    \label{fig:objective}
\end{figure*}

\noindent \textbf{Objectives.} We discuss the objectives presented in Sec.~\ref{subsec:obj} based on our lighting transfer formulation Eqn.~\ref{eqn:lt}. We take $\mathcal{L}_{FR}$ as the baseline, and incorporate other objectives one by one. $\mathcal{L}_{\mathcal{D}\mathcal{R}}$ is used in all the experiments. From Table~\ref{tab:objectives}, there is a remarkable gap between $\mathcal{L}_{L1}$ and $\mathcal{L}_{FR}$. Bringing in $\mathcal{L}_{DSSIM}$ yields a notable improvement over all the metrics. In further, although $\mathcal{L}_{Lab}$ only provides a slight PSNR gain ($+0.34$), it does enhance the lighting effects as reported in Fig.~\ref{fig:objective}. It is worth mentioning that optimizing LighTNet with an auxiliary $\mathcal{L}_{L1}$ loss would largely degrade L1-Norm ($-0.021$) and PSNR ($-0.8$). See Fig.~\ref{fig:objective} for a qualitative evaluation. 
\newline

\noindent \textbf{Image Composition Formulations.} In Table~\ref{tab:decomp-variant} (Top), we study the image composition variants discussed in Sec.~\ref{subsec:rid}.  Overall, our revised formulation $(\mathcal{D} + \mathcal{R} + \alpha) \cdot \mathcal{S}$ outperforms the baseline $\mathcal{D} \cdot \mathcal{S}$ by a significant margin. From the first three columns, while $\mathcal{R}$ supplements reflection effects, the residual $\alpha$ is important in encoding other lighting effects. By investigating $(\mathcal{D} + \mathcal{R} + \alpha) \cdot \mathcal{S}$ \emph{vs.} $(\mathcal{D} + \mathcal{R}) \cdot \mathcal{S} + \alpha$, we find that it would be much better to simulate the PBR compositing process following the product manner. Some qualitative comparisons are shown in Fig.~\ref{fig:decomp}.
\newline

\noindent \textbf{Learning $S'_r$ or Not?} In Eqn.~\ref{eqn:lt}, we choose to learn a residual $S'_r$ to remedy the uneven surfaces of $S'$. There is an alternative that directly estimates a smooth shading $S^{*}$ from $\mathcal{F}_r$. As presented in Table~\ref{tab:decomp-variant} (Bottom), while $(\mathcal{D} + \mathcal{R} + \alpha) \cdot \mathcal{S}^{*}$ improves $(\mathcal{D} + \mathcal{R} + \alpha) \cdot \mathcal{S}'$ by 0.65 on PSNR, our residual architecture significantly yields a PSNR gain of 1.47.
\newline

\begin{figure}[th]
    \centering
    \includegraphics[width=0.48\textwidth]{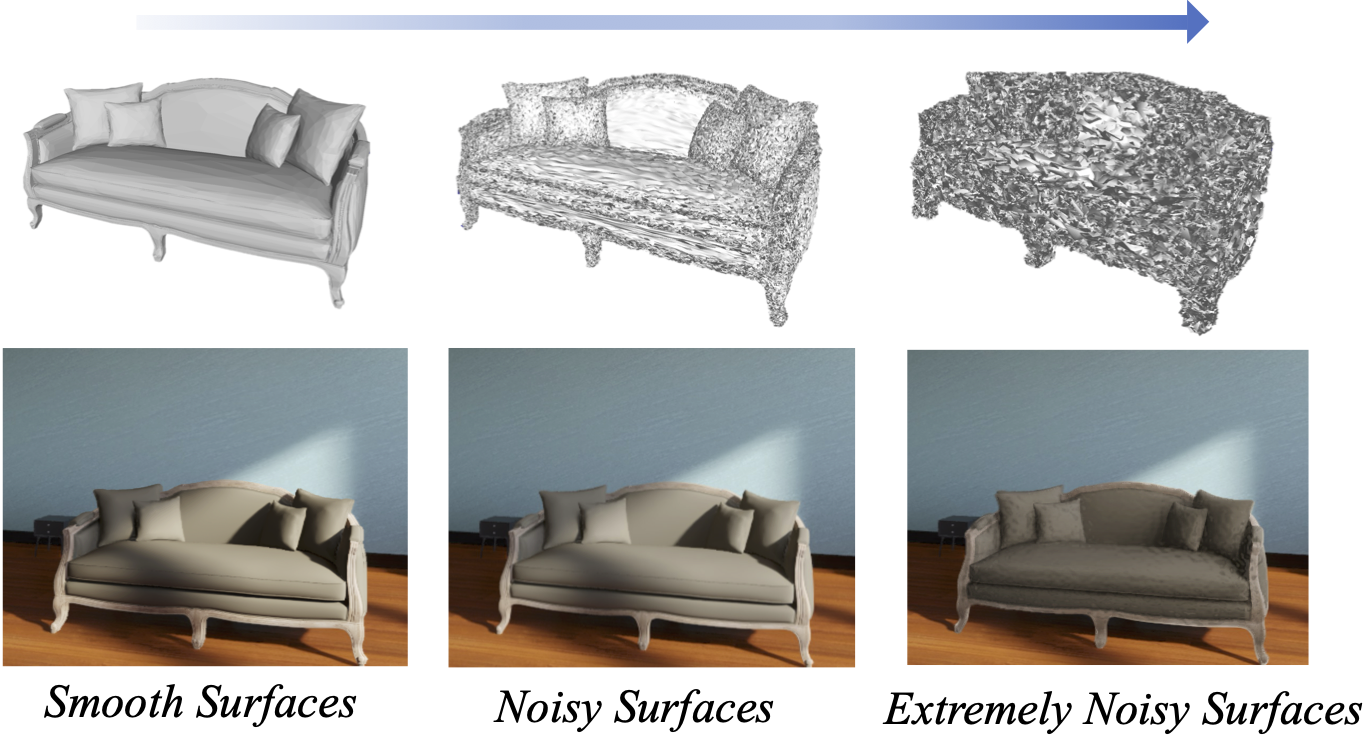}
    \caption{\textbf{Smoothness of R3DMs.} We simulate different levels of noises to the sofa's 3D CAD model. See Sec.~\ref{subsubsec:3df-lighting} for how to add geometry noise to 3DMs.}
    \label{fig:meshes}
\end{figure}

\begin{figure*}[htbp]
    \centering
    \includegraphics[width=0.95\textwidth]{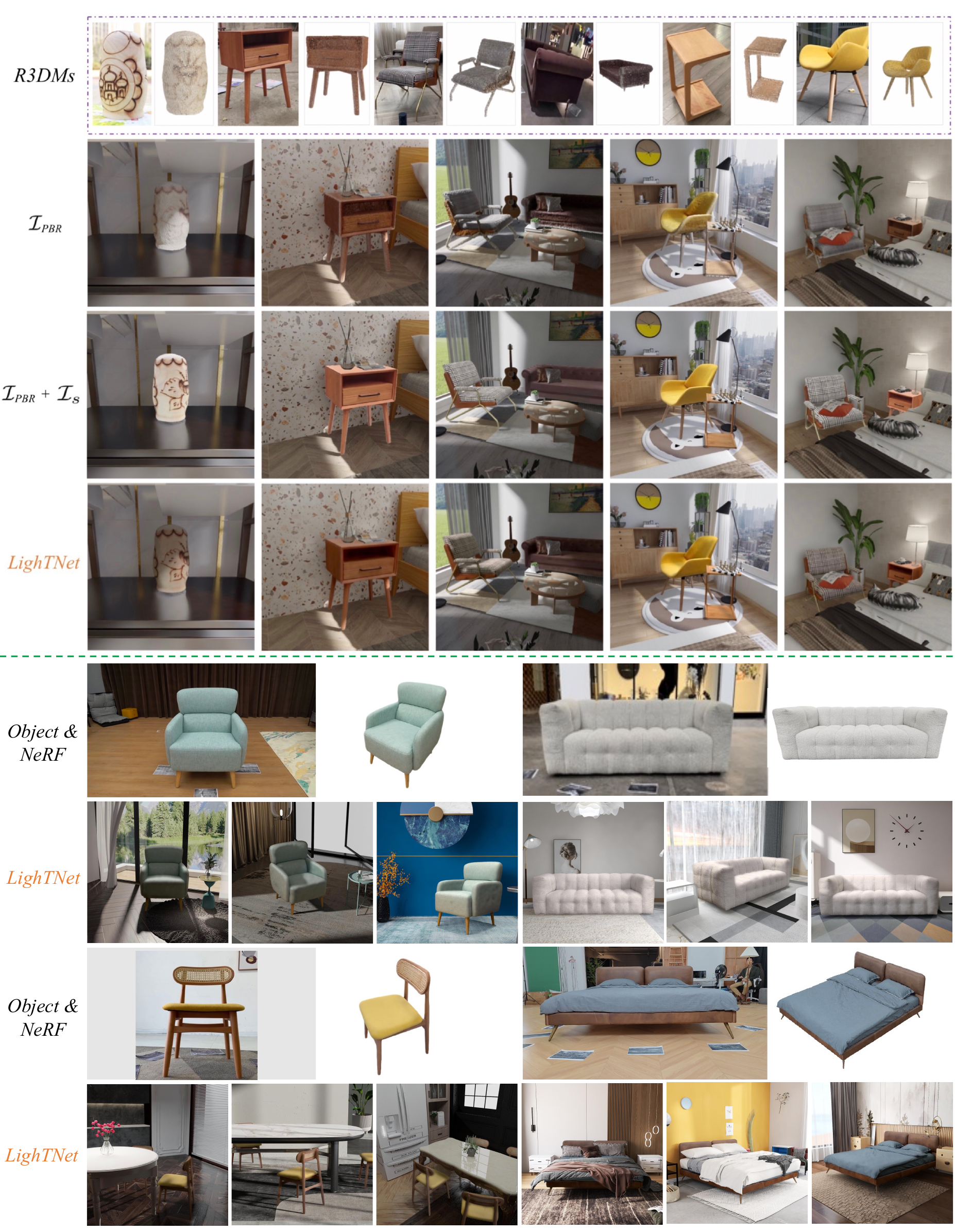}
    \caption{\textbf{Generalizing to Real-Lighting.} {We reconstruct some real objects and use them to create some scenes. See Sec.~\ref{subsec:data-const} for an introduction. \emph{Bottom:} We put the reconstructed objects to different 3D scenes. Here, NeRF means the 2D instance synthesized by NeRF. The lighting details have been successfully preserved by our LighTNet. Please see the shadows caused by object-to-object interactions. 
    Note that, LighTNet here is only trained on the 3DF-Lighting training set. We refer to the supplementary for some rendered videos.}}
    \label{fig:real}
\end{figure*}

\begin{figure*}[htbp]
    \centering
    \includegraphics[width=0.98\textwidth]{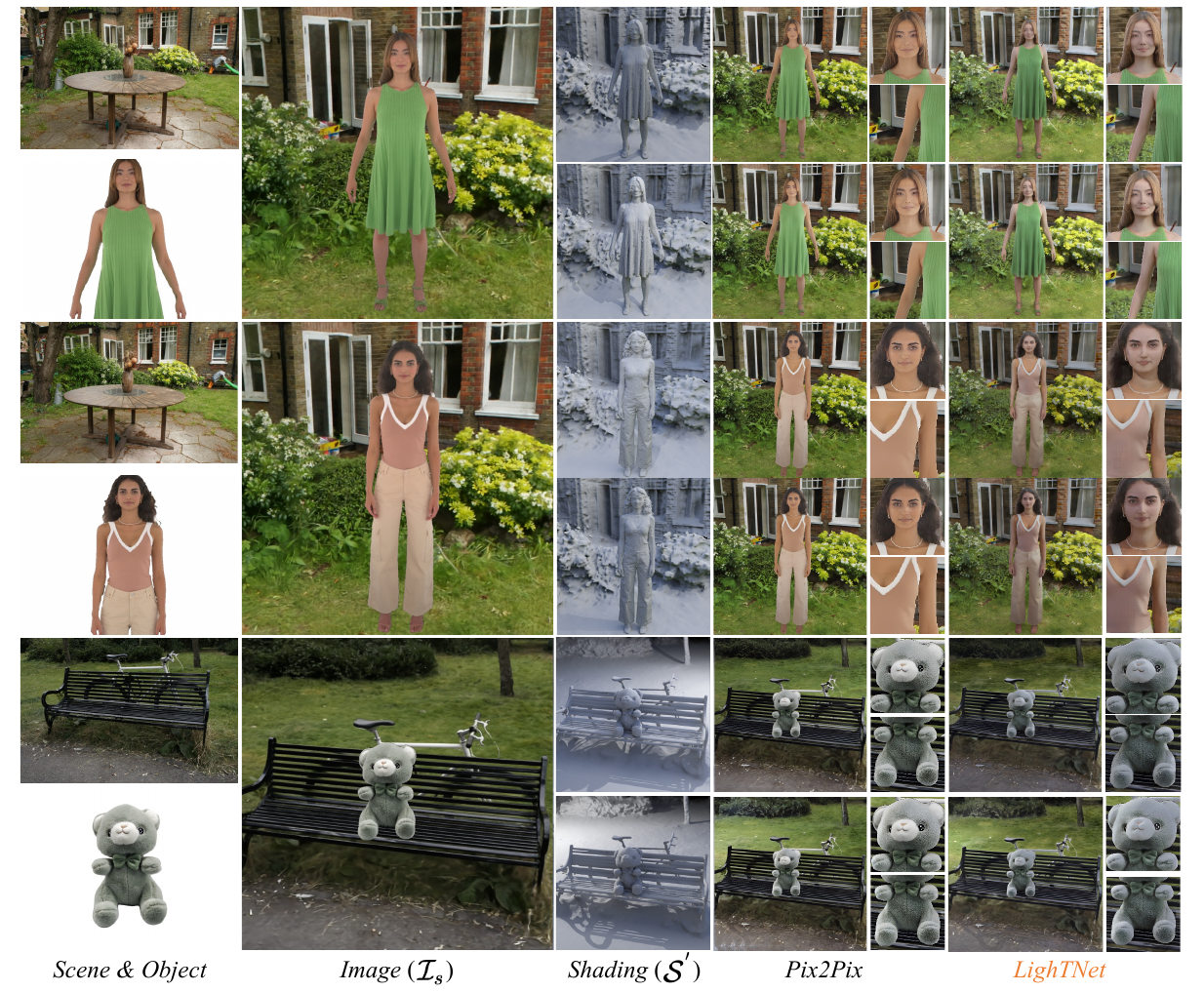}
    \caption{\HF{\textbf{Reconstructed 360 Scenes and Individuals.} We initially reconstruct the scenes, persons, and objects using NeRF approaches and subsequently utilize the reconstructed elements to compose new scenes. For a comprehensive view of the 360 lighting effects, please refer to the supplementary video.}}
    \label{fig:real_scene}
\end{figure*}

\noindent \textbf{Smoothness of R3DMs.} In Fig.~\ref{fig:meshes}, we explore the impact of the smoothness of R3DMs on the final rendering quality. It is not supersizing that if a method can recover smoother surfaces (\emph{e.g.,} NeuS), LighTNet would perform better. For the ``extremely noisy surfaces" case, LighTNet fails to render the lighting details and cannot address the uneven artifacts. The reasons are: (1) a PBR system cannot produce a correct shading map for extremely noisy meshes, as the lighting effects are closely related to surface normals; (2) The capability of LighTNet is not sufficient to remedy these very worse typologies.
\newline

\HF{\noindent \textbf{Weights for Loss Terms.} Here, we simply study $\lambda_1\mathcal{L}_{FR} + \lambda_2\mathcal{L}_{DSSIM} + \lambda_3\mathcal{L}_{Lab}$ through a controlled variable technique. The scores are reported in Fig. \ref{fig:loss_weights}. For efficiency, this ablation was performed on a reduced subset (1/5) of 3DF-Lighting.}

\setlength\tabcolsep{3pt}
\begin{table}[t!]
\caption{\textbf{Image Composition Formulations.} See Sec.~\ref{subsec:rid} and Sec.~\ref{subsec:arch} for explanations of these formulations. $\mathcal{S}^{*}$ means that we directly predict a smooth shading from $\mathcal{F}_r$ instead of estimating the shading residual $\mathcal{S}'_r$.}
\centering
\footnotesize
\begin{tabular}{ c  c  c  c  c }
\toprule
Variant & \emph{L1}-Norm $\downarrow$ & PSNR $\uparrow$ & SSIM $\uparrow$ & \emph{Lab} Angle $\downarrow$ \\
\cmidrule(lr){1-1}\cmidrule(lr){2-5}
\multicolumn{5}{c}{Composition Formulations ($\mathcal{S} = \mathcal{S}' + \mathcal{S}'_r$)} \\
\midrule\midrule
$\mathcal{D} \cdot \mathcal{S}$ &0.0283 &28.46 &0.9128 & 0.3449 \\
$(\mathcal{D} + \mathcal{R}) \cdot \mathcal{S}$ &0.0262 &28.83 &0.9117 & 0.3411 \\
$(\mathcal{D} + \mathcal{R}) \cdot \mathcal{S} + \alpha$ &0.0238 &29.45 &0.9040 & 0.3458 \\
$(\mathcal{D} + \mathcal{R} + \alpha) \cdot \mathcal{S}$ & \textbf{0.0219} & \textbf{30.17} & \textbf{0.9142} & \textbf{0.3137} \\
\midrule
\multicolumn{5}{c}{Architecture: Learning $\mathcal{S}'_r$ or Not} \\
\midrule\midrule
$(\mathcal{D} + \mathcal{R} + \alpha) \cdot \mathcal{S}'$ &0.0260 &28.70 &0.8935 & 0.3574 \\
$(\mathcal{D} + \mathcal{R} + \alpha) \cdot \mathcal{S}^{*}$ &0.0241 &29.35 &0.8996 & 0.3473 \\
$(\mathcal{D} + \mathcal{R} + \alpha) \cdot (\mathcal{S}' + \mathcal{S}'_r)$ & \textbf{0.0219} & \textbf{30.17} & \textbf{0.9142} & \textbf{0.3137} \\
\bottomrule
\end{tabular}
\label{tab:decomp-variant}
\end{table}

\section{Discussion}
\noindent \textbf{Setting.} \HF{Our primary insight of develop such a lighting transfer avenue is inspired by the industry production pipeline. For example, in order to showcase furniture online, a furniture seller typically needs artists to reconstruct 3D CAD models of their furniture. Subsequently, designers use these 3D furniture models to create a virtual 3D CAD scene, and a rendering engine is employed to generate images and videos that effectively showcase the furniture. Crafting high-quality 3D furniture models is a costly endeavor, especially when dealing with unique materials or styles. Moreover, achieving high-quality furniture reconstructions requires the expertise of professional artists.  With the motivation,} we focus on how to flexibly integrate automatically reconstructed 3D models into practical 3D modeling workflows. \HF{Essentially, with the proposed lighting transfer avenue, one can import newly reconstructed Rough 3D Models (R3DMs) into graphical software (e.g., 3DS-Max). This allows users to design diverse indoor and outdoor 3D scenes and collaboratively render content alongside other high-quality 3D CAD models (3DMs).} \HF{It's noteworthy that NeRF in our paper serves as a specific example to illustrate the potential of the lighting transfer avenue. In our approach, the R3DM component is independent of the Neural Fields Rendering (NFR) synthesis (or NeRF) segment. For a real object, we can employ one method to reconstruct its R3DM and another method for 2D instance synthesis if it outperforms NeRF in novel view synthesis.}
\newline

\noindent \textbf{Relation to Inverse Rending with Implicit Neural Representation.}
Leveraging implicit neural representation, recent inverse rendering works can decompose a scene under complex and unknown illumination into spatially varying BRDF material properties \cite{physg2020,zhang2021nerfactor,boss2020nerd,srinivasan2020nerv,zhang2022modeling}. These techniques enable material editing and free view relighting of the reconstructed scene. Here, we take NeRFactor \cite{nerfactor} as an example to discuss the main differences of these works and the raised lighting transfer avenue. First, NeRFactor focused on estimate SVBRDF properties of single scene or object. Beyond free-view relighting, we can imagine that NeRFactor supports the object inserting application, \emph{i.e.,} inserting a 3D object into a \emph{static image}, as we can extract the global lighting probes from the target static image. But if we would like to insert multiple 3D objects into a single image, NeRFactor would overlook the possible indirect lighting effects caused by object-to-object occlusion because there is no a ``3D scene" concept involved. It's worth mentioning NeRFactor can simulate local shadows caused by self-occlusion of its reconstructed single object. In contrast, we study a more practical problem that is ``can we use rough 3D models, together with 3D CAD models drawing by artists, to create arbitrary \emph{3D scenes} and render high-quality contents?". Thereby, our studied setting is totally different from the setting of NeRFactor. Second, in the proposed lighting transfer avenue, the R3DM part is independent with the NFR synthesis part. For a real object, we can use a method to reconstruct its R3DM, and use another method to perform 2D instance synthesizing. This paper takes NeRF as an example to explain the proposed avenue because: (1) it supports high-quality novel view synthesis; and (2) at the same time, we can conveniently extract a R3DM from a trained NeRF via a marching cube algorithm. From this perspective, we can apply NeRFactor instead of NeRF for 2D instance synthesizing. But as analyzed before, how should we handle the indirect lighting details caused by the interaction of multiple 3D objects? The introduced LighTNet could give a possible answer to this question.

\section{Limitation}
\label{sec:limitation}
\begin{figure}[t!]
    \centering
    \includegraphics[width=0.48\textwidth]{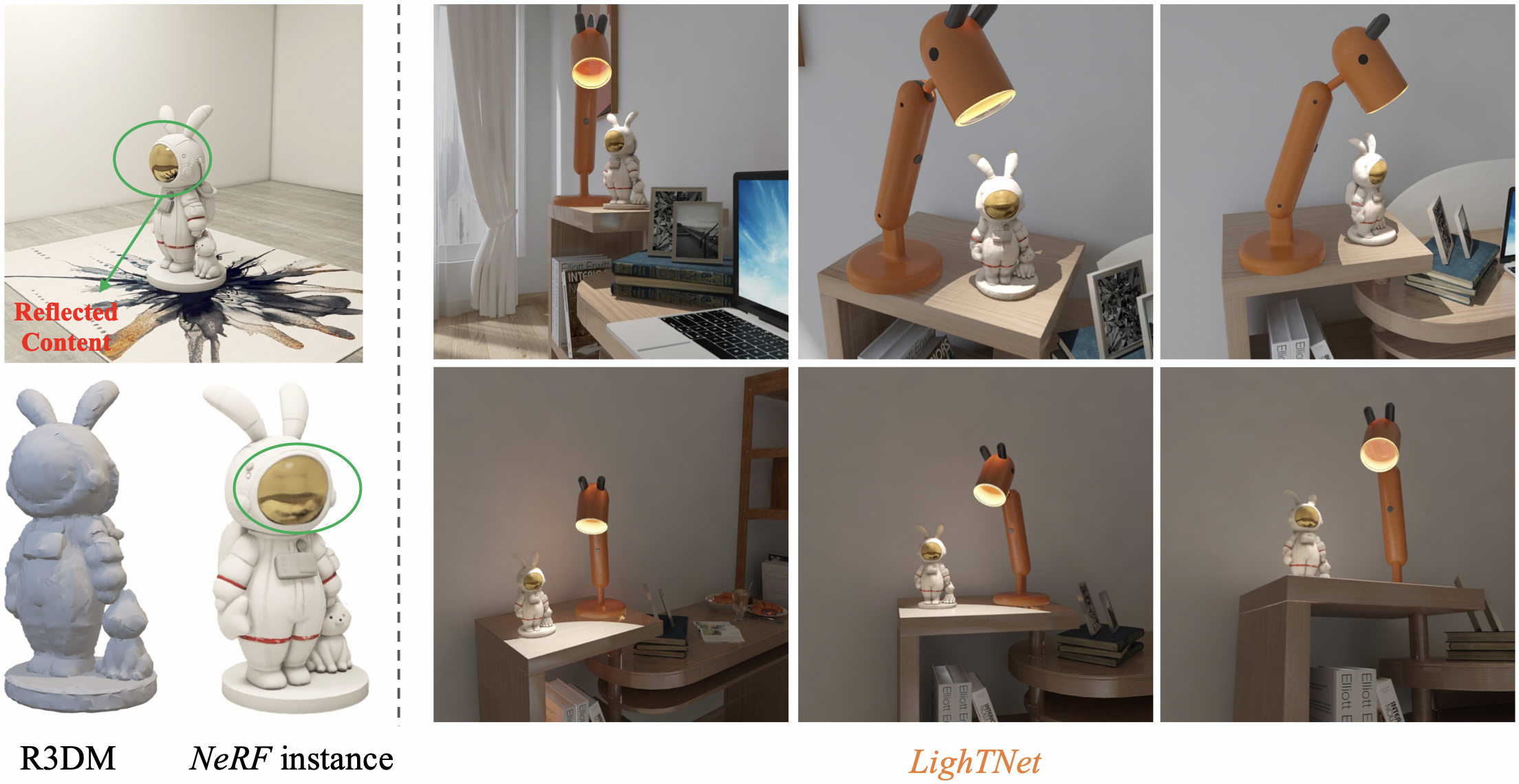}
    \caption{{\textbf{Failure Case.} One of the major limitations of LighTNet is it cannot handle strong specular materials. The reflected content would be incorrectly maintained during the rendering process. Zoom in for a better view.
    }}
    \label{fig:limitation}
\end{figure}

\noindent \textbf{Tough Materials.} LighTNet cannot handle strong specular materials yet. As shown in Fig.~\ref{fig:limitation}, the synthesized 2D instances by NeRF contain the reflected content. It's unavoidable yet as NeRF series learn to fit a \emph{captured scene} for its free view synthesis. We find LighTNet would preserve the reflected content while ignoring the real reflected content of the newly created scenes. A possible reason is that we do not have strong specular materials in our training set, as 3D-FRONT only shared several specular objects. It should be one of the major limitations. This issue would disappear if future NeRF research or other view synthesis works can disentangle the reflected content from the objects' textures. \HF{Moreover, our approach struggles with scattering materials (e.g., clouds) and objects with intricate structures (e.g., Eiffel Tower). The current methodology falls short in recovering the meaningful geometry of these categories, resulting in a "Shading" map riddled with considerable noise.}
\newline

\HF{\noindent \textbf{Blurriness \& Dark spots.} Our renderings exhibit slight blurriness compared to ground truth (GT) images and contain artifacts such as dark spots. Regarding the blurriness, it's reasonable to expect that images generated by a standard CNN, like our composition network, may not seamlessly match the quality of those rendered by advanced rendering engines. As for the artifacts, it's possible that the "shading residual" approach, while performing better than the compared image composition variants in Fig. \textcolor{red}{7}, may not fully address the issue of uneven surfaces. We suggest that further investigation into techniques in super-resolution and diffusion models could address these challenges.}
\newline

\HF{\noindent \textbf{Rendering Speed.} Presently, rendering an image using our method on a V100 GPU takes approximately 4 seconds. Thereby, our approach only support off-line rendering at this time. As a complement, training LighTNet on 3DF-Lighting requires approximately 12 hours utilizing a single V100 GPU. To speed up the rendering process, the UNet structure might be deeply revised and optimized.}
\newline

\HF{\noindent \textbf{Stability for Video Rendering.} 
Another limitation is that our approach currently encounters stability issues in video rendering, as we have focused solely on rendering static images in this paper. To quantitatively evaluate stability, we constructed a 360-degree test set using six furniture shapes and twelve 3D scenes from 3D-FRONT. Initially, we pre-defined a light source and energy for each scene, randomly rendering 200 images from viewpoints sampled on a full sphere to train NeRF and extract its R3DM. Subsequently, we placed the object's R3DM into the scene, generated a random light source and energy for each 3D scene, and rendered 100 $\mathcal{S}'$ and $\mathcal{M}$ from viewpoints uniformly sampled on a semi-sphere around the object's R3DM. Simultaneously, we synthesized the corresponding 100 $\mathcal{I}_s$ using its NeRF. Finally, we rendered the 100 $\bar{\mathcal{I}}_t$ (ground-truth images) at size 800 $\times$ 800 by replacing the R3DM with its 3DM. Through this workflow, we constructed a 360-degree test set with 600 samples ${(\mathcal{I}_s, \mathcal{S}', \bar{\mathcal{I}}_t, \mathcal{M})}$. As reported in Table~\ref{tab:eval360scene}, although LighTNet outperforms Pix2Pix~\cite{isola2017image} in all metrics, the variances of \emph{L1}-Norm and PSNR are larger than Pix2Pix~\cite{isola2017image}. It makes sense since the final rendering of Pix2Pix consistently contains weak lighting effects. Modeling the temporal consistency for LighTNet would be a promising avenue for future research to overcome this limitation.}

\setlength\tabcolsep{4pt}
\begin{table}[t!]
\caption{\HF{\textbf{Stability for Video Rendering.}} }
\centering
\small
\begin{tabular}{ c  c  c }
\toprule
Metric & Pix2Pix\cite{isola2017image} & LighTNet \\
\cmidrule(lr){1-1}\cmidrule(lr){2-3}
\emph{L1}-Norm $\downarrow$ & 0.0366 $\pm$ 0.0081 & 0.0275 $\pm$ 0.0097 \\
PSNR $\uparrow$ & 27.5716 $\pm$ 1.4850 & 31.7297 $\pm$ 1.9355 \\
SSIM $\uparrow$ & 0.9167 $\pm$ 0.0168 & 0.9356 $\pm$ 0.0103 \\
\emph{Lab} Angle $\downarrow$ & 0.4503 $\pm$ 0.0197 & 0.3275 $\pm$ 0.0114 \\
\bottomrule
\end{tabular}
\label{tab:eval360scene}
\end{table}

\section{Conclusion}
 In this paper, we are prudent to rethink reconstructed rough 3D models (R3DMs) and present a lighting transfer avenue to flexibly integrate R3DMs into practical 3D modeling workflows such as 3D scene creation, lighting editing, and rendering. Physically-based rendering (PBR) would render low-quality images of scenes constructed by R3DMs. A remedy is to represent real-world objects as individual neural fields (e.g. NeRF) in addition to R3DMs, as neural fields rendering (NFR) can synthesize photo-realistic object images under desired viewpoints. The main question is that NFR instances cannot reflect the lighting details on R3DMs. We thus present a lighting transfer network (LighTNet) as a solution. LighTNet reasons about a reformulated image composition model and can bridge the lighting gaps between NFR and PBR, such that they can benefit from each other. Moreover, we introduce a new \emph{Lab} angle loss to enhance the contrast between lighting strength and colors. Qualitative and quantitative comparisons show the superiority of LighTNet in preserving both direct and indirect lighting effects.


%

\ifCLASSOPTIONcaptionsoff
  \newpage
\fi



%

\bibliographystyle{IEEEtran}
\bibliography{egbib_abbr}


%

\begin{IEEEbiography}[{\includegraphics[width=1in,height=1.25in,clip]{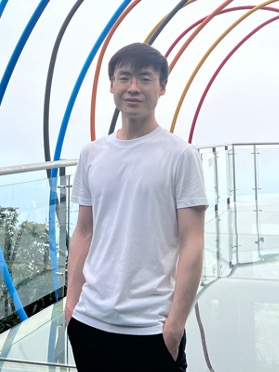}}]{Bowen Cai}
is an Algorithm Expert at Tao Technology Department, Alibaba Group. He received his PhD degree from Beihang University. After that, he was a postdoctoral researcher, working with Prof. Hua Li at the Institute of Computing Technology, Chinese Academy of Sciences and Yinghui Xu at Alibaba Group. His interests include semantic segmentation, 3D reconstruction, and neural rendering.
\end{IEEEbiography}

\begin{IEEEbiography}[{\includegraphics[width=1in,height=1.25in,clip]{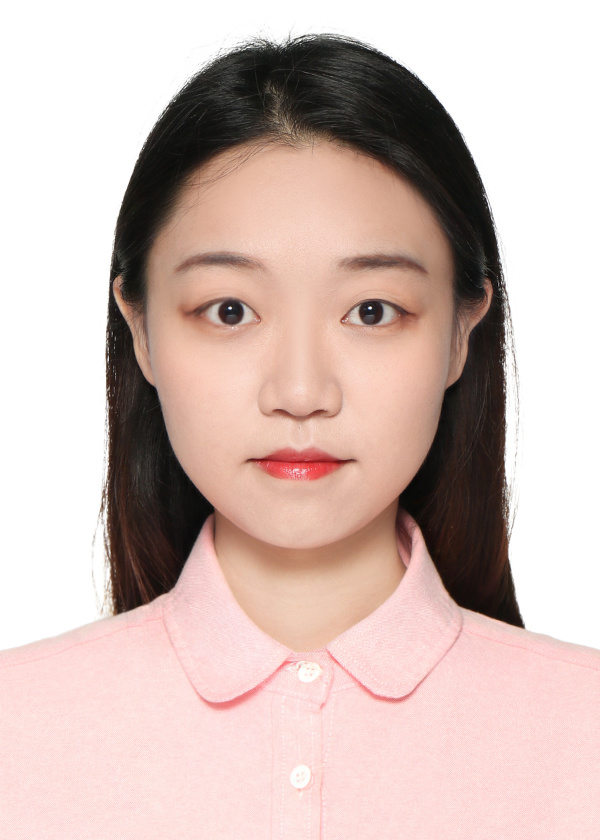}}]{Yujie Li}
was an Algorithm Engineer at Tao Technology Department, Alibaba Group. She received her master degree from Beijing University of Technology, in 2020. She currently works on 3D scene synthesis and neural rendering.
\end{IEEEbiography}

\begin{IEEEbiography}[{\includegraphics[width=1in,height=1.25in,clip]{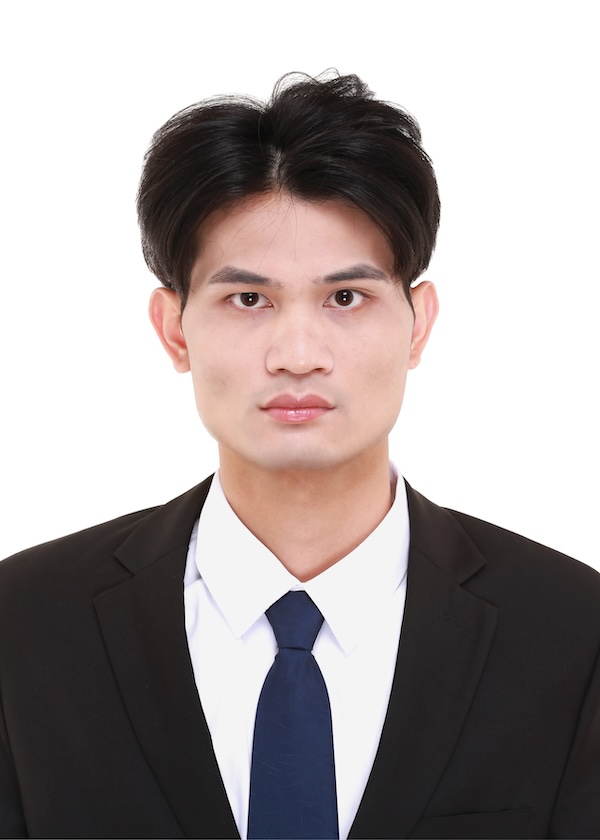}}]{Yuqing Liang}
is an Algorithm Engineer at Tao Technology Department, Alibaba Group. He received her master degree from Najing University, in 2022. He currently works on 3D reconstruction and neural rendering.
\end{IEEEbiography}

\begin{IEEEbiography}[{\includegraphics[width=1in,height=1.25in,clip]{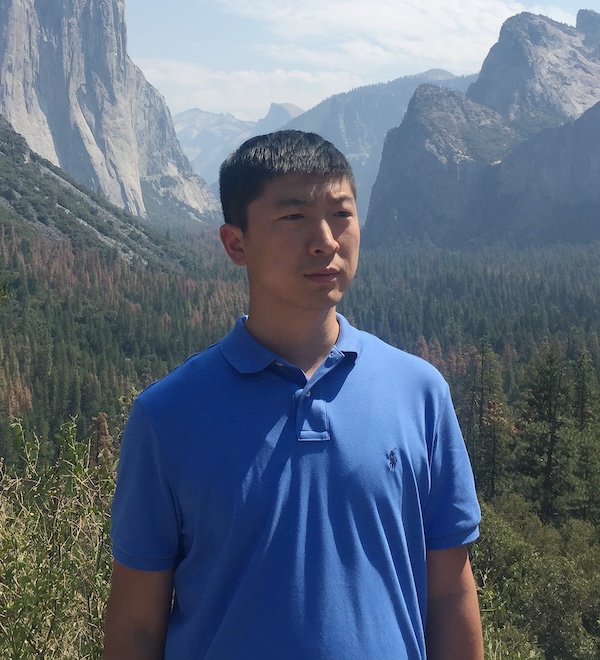}}]{Rongfei Jia}
is a Staff Algorithm Expert at Tao Technology Department, Alibaba Group. He received his PhD degree from Beihang University. He is leading an algorithm team which is devoted in 3D AI modeling and generation technologies. His research interests include 3D object modeling, scene understanding and synthesis.
\end{IEEEbiography}

\begin{IEEEbiography}[{\includegraphics[width=1in,height=1.25in,clip]{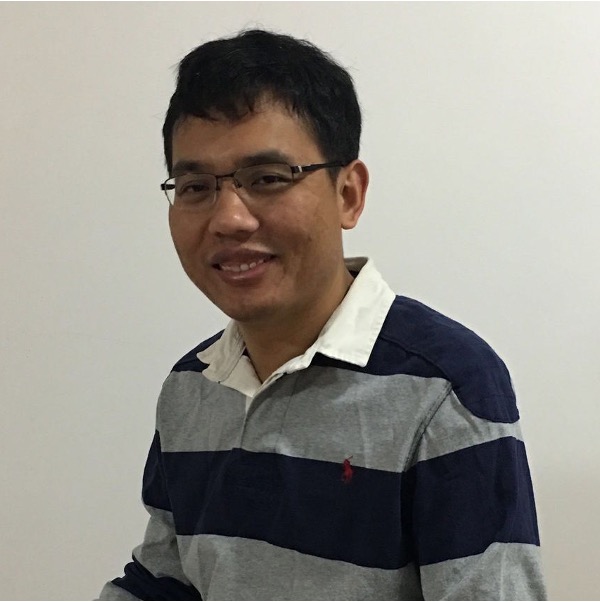}}]{Bingqiang Zhao}
was a Senior Staff Algorithm Expert at Tao Technology Department, Alibaba Group. He  graduated  from  Tsinghua  University  in  2006. He  was the head of Taobao Business Machine Intelligence Department. 
\end{IEEEbiography}

\begin{IEEEbiography}[{\includegraphics[width=1in,height=1.25in,clip]{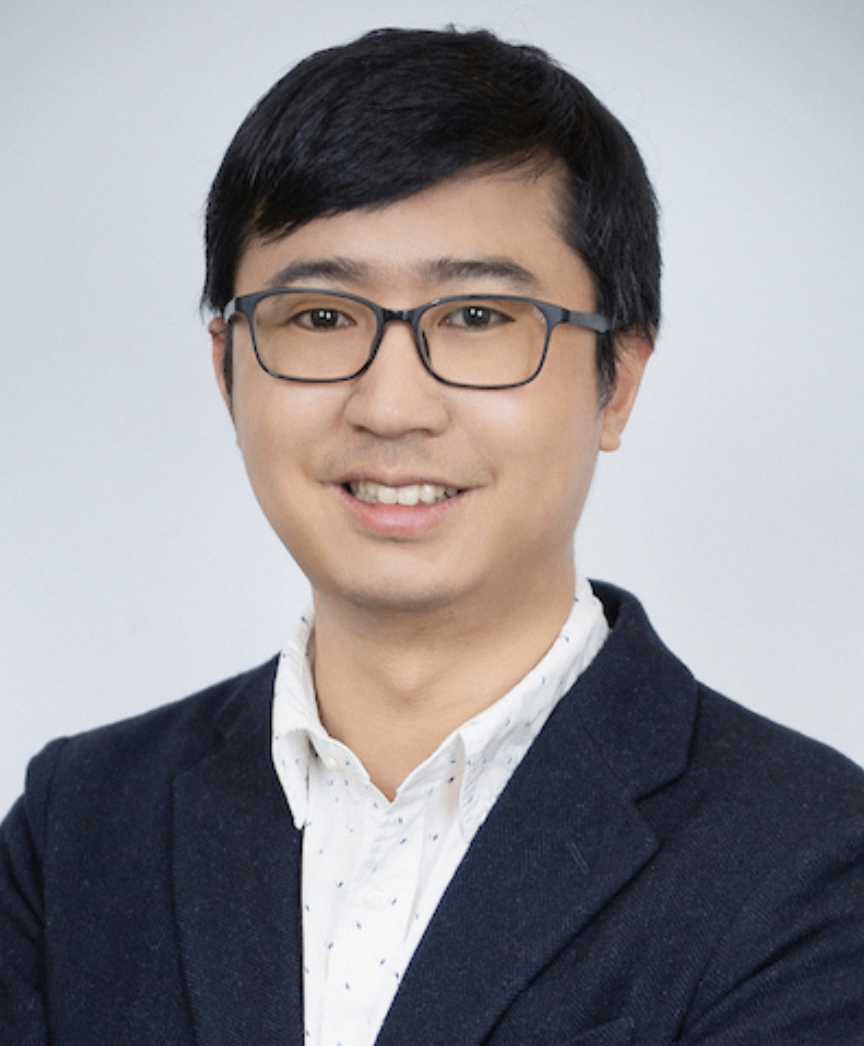}}]{Mingming Gong}
is a Senior Lecturer in data science with the School of Mathematics and Statistics, the University of Melbourne. He has authored and co-authored 60+ research papers on top venues such as ICML, NeurIPS, ICLR, CVPR, TPAMI, and IJCV with 10+ oral/spotlight presentations. He received the Discovery Early Career Researcher Award from Australian Research Council in 2021. His research interests include causal reasoning, machine learning, and computer vision.
\end{IEEEbiography}

\begin{IEEEbiography}[{\includegraphics[width=1in,height=1.25in,clip]{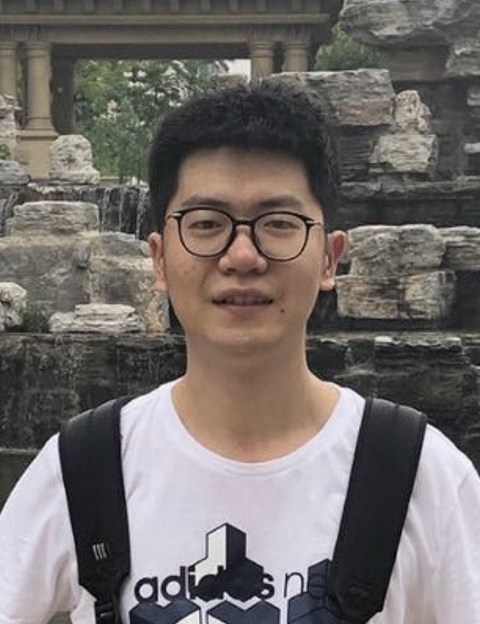}}]{Huan Fu}
is a Staff Algorithm Expert at Tao Technology Department, Alibaba Group. He received the BSc degree from University of Science and Technology of China, in 2014, and the PhD degree from The University of Sydney, Australia, in 2019.  He has authored and co-authored 20+ research papers on top venues such as NeurIPS, CVPR, ICCV, IJCV, and TPAMI, with a best paper finalist in CVPR19. His research interests include 2D/3D scene understanding, 3D reconstruction, and neural rendering.
\end{IEEEbiography}





\clearpage

\beginsupplement

\section*{Supplementary Material}

\section*{Contents}
The supplementary materials consist of:
\begin{itemize}
  \item The detailed network architectures.
  \item Two videos that records that an artist designs some rooms via R3DMs and R3Ds and renders high-quality images or videos leveraging LighTNet.
  \item More qualitative results (images and videos).
\end{itemize}

\begin{figure*}[th!]
    \centering
    \includegraphics[width=0.98\textwidth]{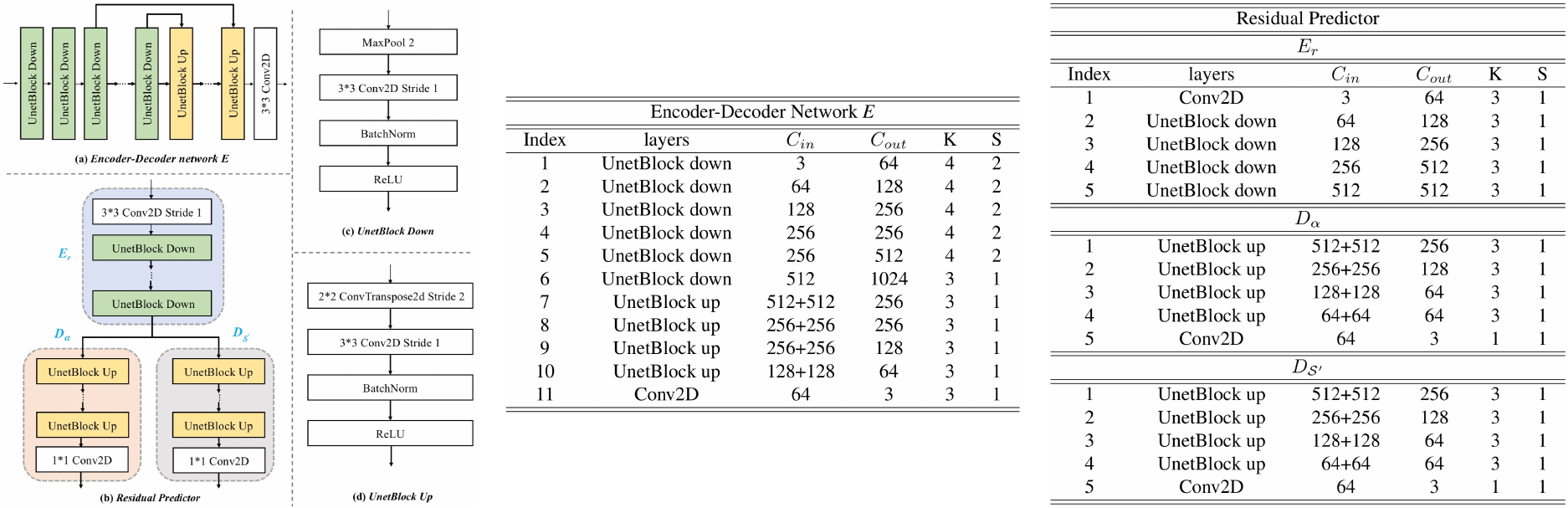}
    \caption{{\textbf{Network Architecture.}} (a). Encoder-Decoder Network \emph{E}. (b). Residual Predictor. $D_\alpha$ and $D_{\mathcal{S}'}$ share the same encoder $E_r$, and are used to estimate the residual lighting effect $\alpha$ and shading $\mathcal{S}'$, respectively. (c). The U-Net downsampling block in \emph{E} and $E_r$. (d). The U-Net upsampling block in \emph{E}, $D_\alpha$ and $D_{\mathcal{S}'}$. Zoom in for a better view.}
    \label{subfig:LighTNet}
\end{figure*}

\section{Network Architectures}

The network architectures for the lighting transfer network (LighTNet) are reported in S.Figure~\ref{subfig:LighTNet}. For convenience, we use the following abbreviation: $C_{in}$ = Input Channel, $C_{out}$ = Feature Channel, K = Kernel Size, S = Stride Size, Conv2D = Convolutional Layer. 

\section{More Qualitative Results}
\begin{figure*}[h]
    \centering
    \setlength{\abovecaptionskip}{0cm}
    \includegraphics[width=0.76\textwidth]{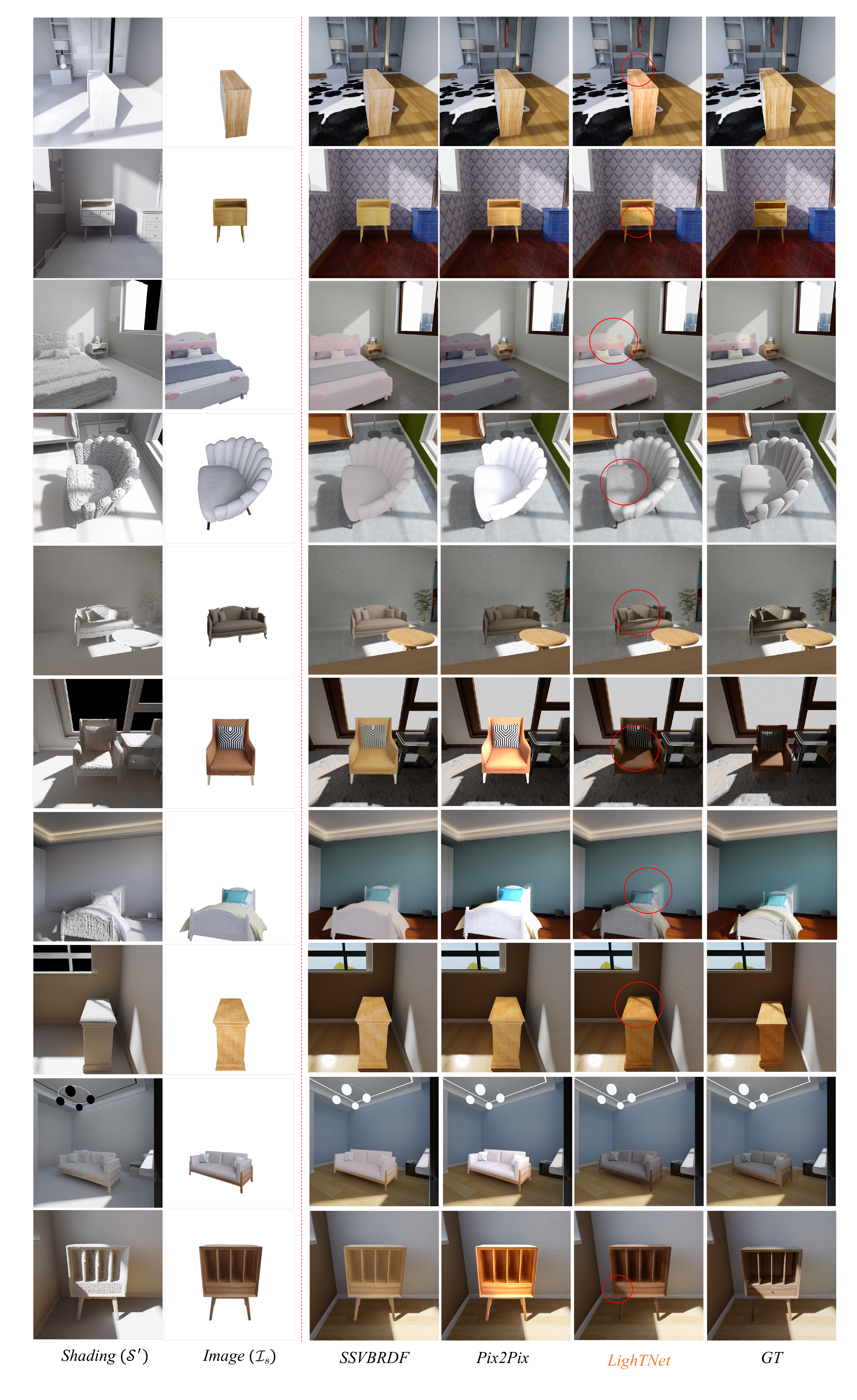}
    \caption{{More qualitative comparisons with baseline methods on the 3DF-Lighting test set. LighTNet can well preserve the lighting details (\emph{e.g.,} local shadows.)} Zoom in for a better view. }
    \label{supfig:comparison}
\end{figure*}

\begin{figure*}[t!]
    \centering
    \includegraphics[width=0.98\textwidth]{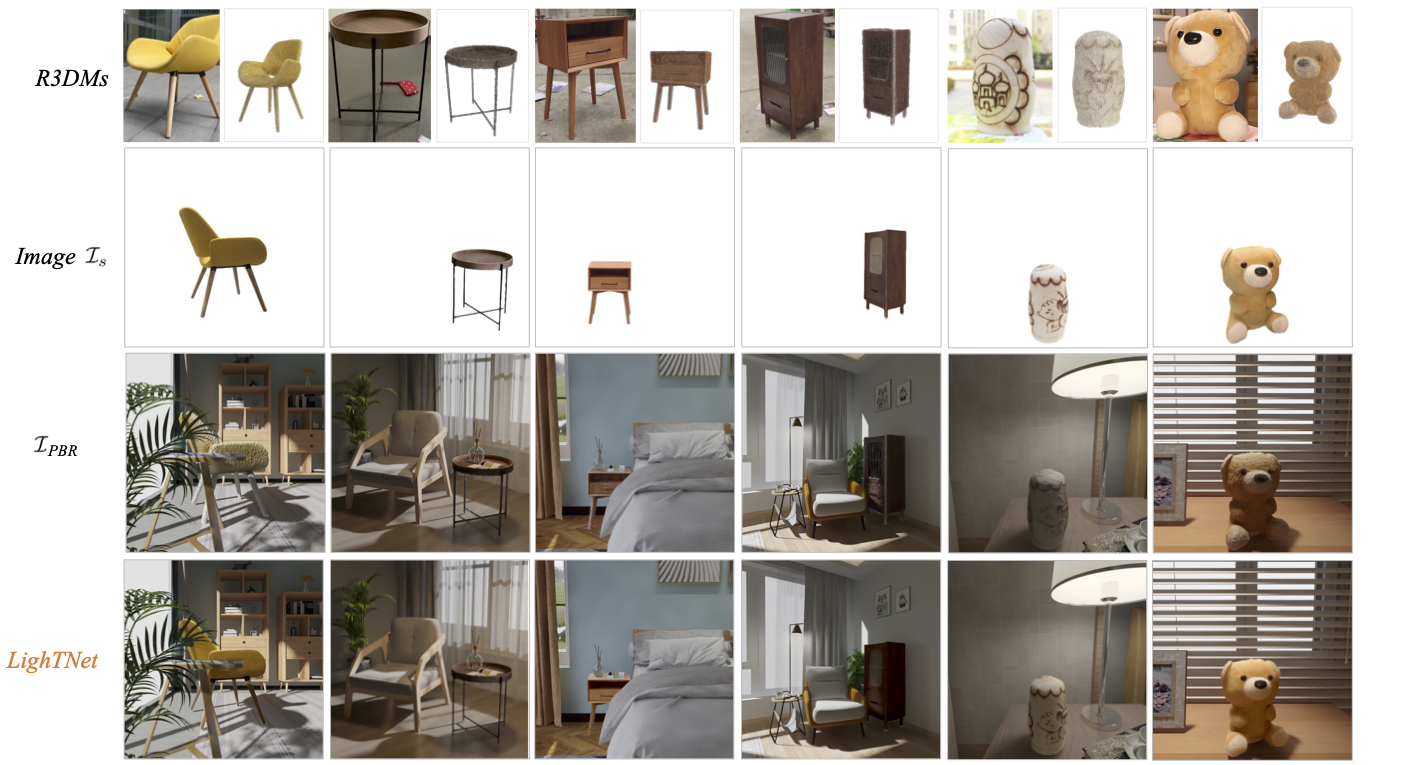}
    \caption{ More rendered results of scenes created by R3DMs of real objects. Zoom in for a better view. }
    \label{supfig:real_lighting}

\end{figure*}

\begin{figure*}[t!]
    \centering
    \includegraphics[width=0.88\textwidth]{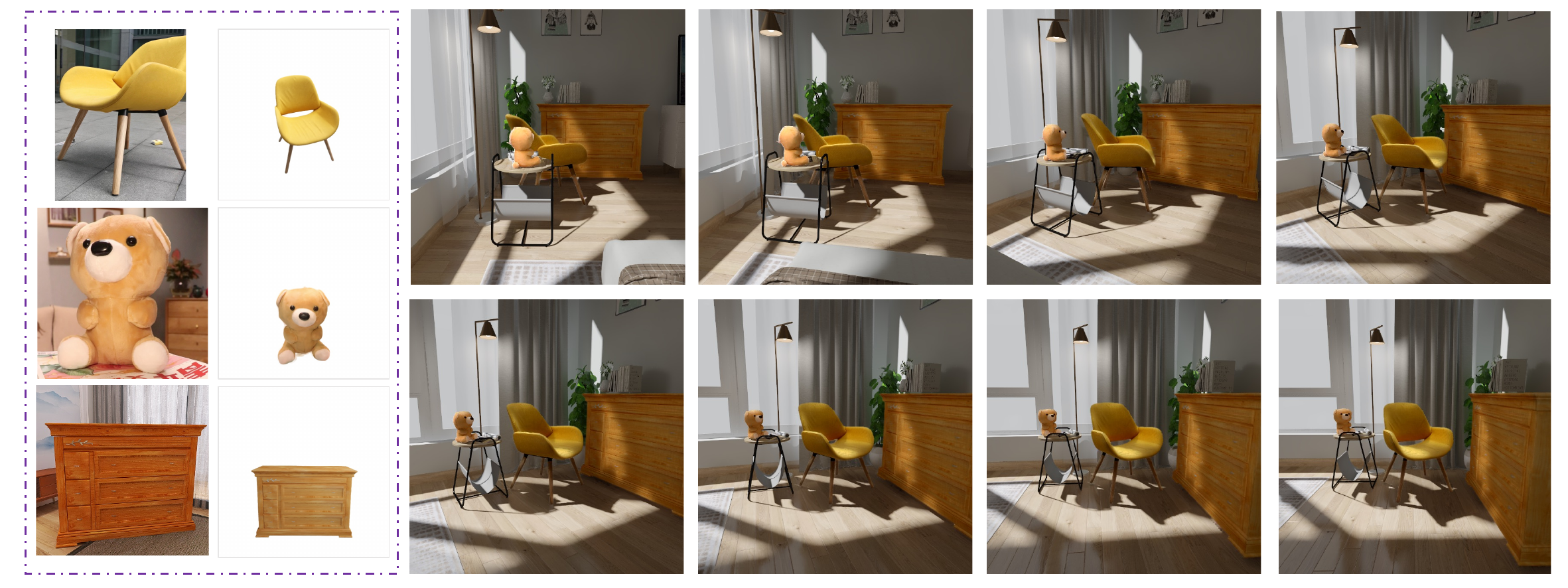}
    \caption{ The R3DMs including bear, chair, and cabinet are put into room to make it collaborate with other 3D CAD furniture. Their lightings are transferred by LighTNet, while other 3D CAD models are rendered by PBR. We can see that R3DMs transferred by LighTNet can be well compatible with the PBR scene. Zoom in for a better view. }
    \label{supfig:real_lighting_v2}
\end{figure*}

\begin{figure*}[t!]
    \centering
    \includegraphics[width=0.88\textwidth]{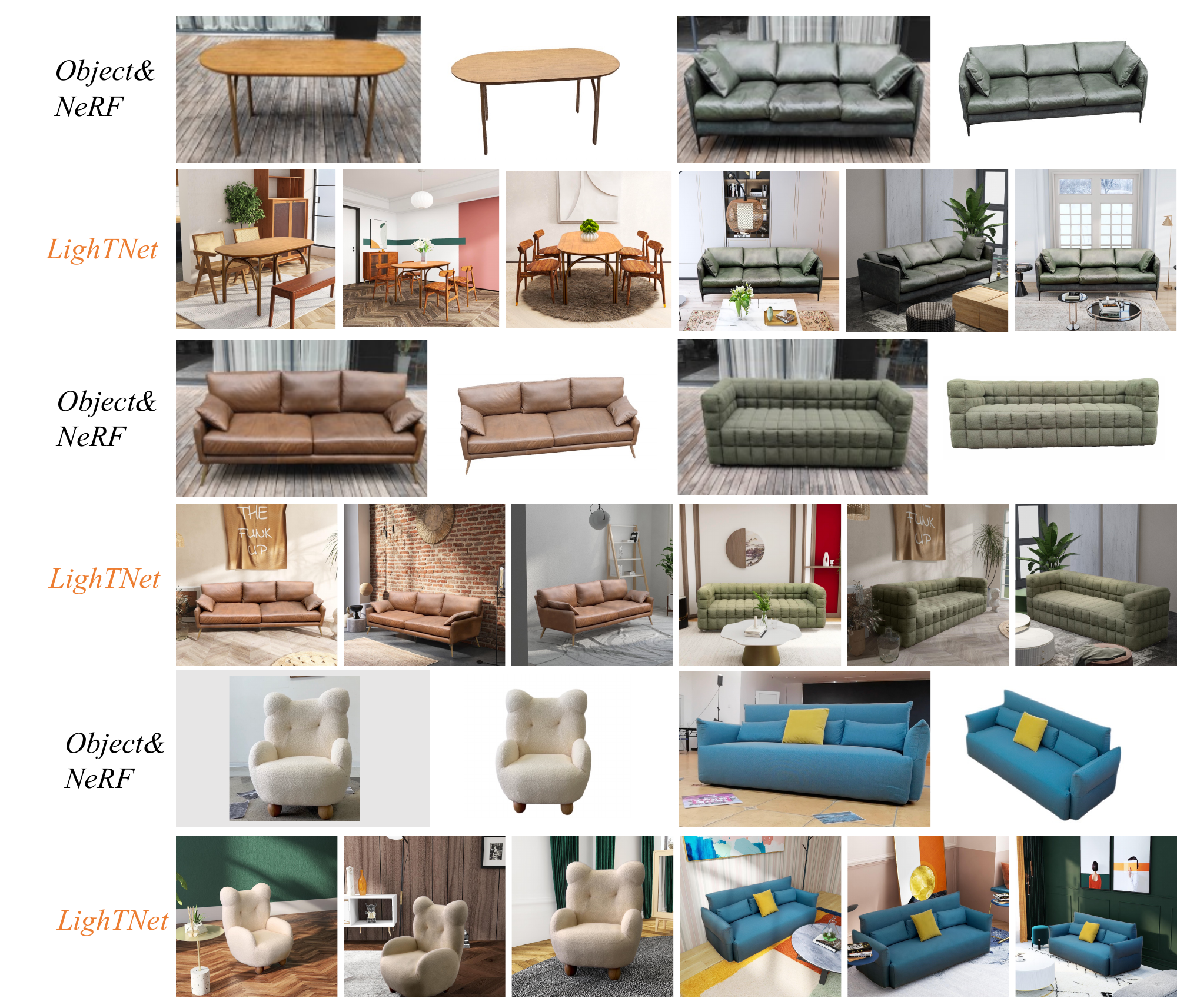}
    \caption{ We put the reconstructed objects to different 3D scenes. Here, NeRF means the 2D instance synthesized by NeRF. The lighting details have been successfully preserved by our LighTNet approach.}
    \label{supfig:more_obj}
\end{figure*}

S.Figure~\ref{supfig:comparison} provides more qualitative comparisons on the 3DF-Lighting test set. In S.Figure~\ref{supfig:real_lighting}-\ref{supfig:more_obj} and the supplemental video, we incorporate more rendered results of scenes created by R3DMs of real objects (See "Generalizing to Real-Lighting" in the main paper).

\end{document}